\documentclass[10pt,twocolumn,letterpaper]{article}

\usepackage[pagenumbers]{wacv} % To force page numbers, e.g. for an arXiv version

\usepackage{graphicx}
\usepackage{amsmath}
\usepackage{amssymb}
\usepackage{booktabs}

\usepackage[pagebackref,breaklinks,colorlinks]{hyperref}

\usepackage[capitalize]{cleveref}
\crefname{section}{Sec.}{Secs.}
\Crefname{section}{Section}{Sections}
\Crefname{table}{Table}{Tables}
\crefname{table}{Tab.}{Tabs.}

\def\wacvPaperID{1894} % *** Enter the WACV Paper ID here
\def\confName{WACV}
\def\confYear{2025}

\begin{document}


\title{Skyeyes: Ground Roaming using Aerial View Images}

\author{
Zhiyuan Gao\textsuperscript{1,2,\thanks{Equal Contribution}}\quad Wenbin Teng\textsuperscript{1,2,\footnotemark[1]}\quad Gonglin Chen\textsuperscript{1,2}\quad Jinsen Wu\textsuperscript{1,2}\\
Ningli Xu\textsuperscript{3} \quad Rongjun Qin\textsuperscript{3} \quad Andrew Feng\textsuperscript{2} \quad Yajie Zhao\textsuperscript{1,2,\thanks{Corresponding Author}}\\\\
\textsuperscript{1}University of Southern California\quad \textsuperscript{2}Institute for Creative Technologies\quad \textsuperscript{3}The Ohio State University\\
\ttfamily\small \{gaozhiyu, wenbinte, gonglinc, jinsenwu\}@usc.edu\\
\ttfamily\small \{xu.3961\}@buckeyemail.osu.edu \quad \{Qin.324\}@osu.edu \quad \{feng, zhao\}@ict.usc.edu
}

\date{}

\vspace{-1cm}

\makeatletter
\let\@oldmaketitle\@maketitle% Store \@maketitle
\renewcommand{\@maketitle}{\@oldmaketitle% Update \@maketitle to insert...
  \centering
  \includegraphics[width=0.95\linewidth]
    {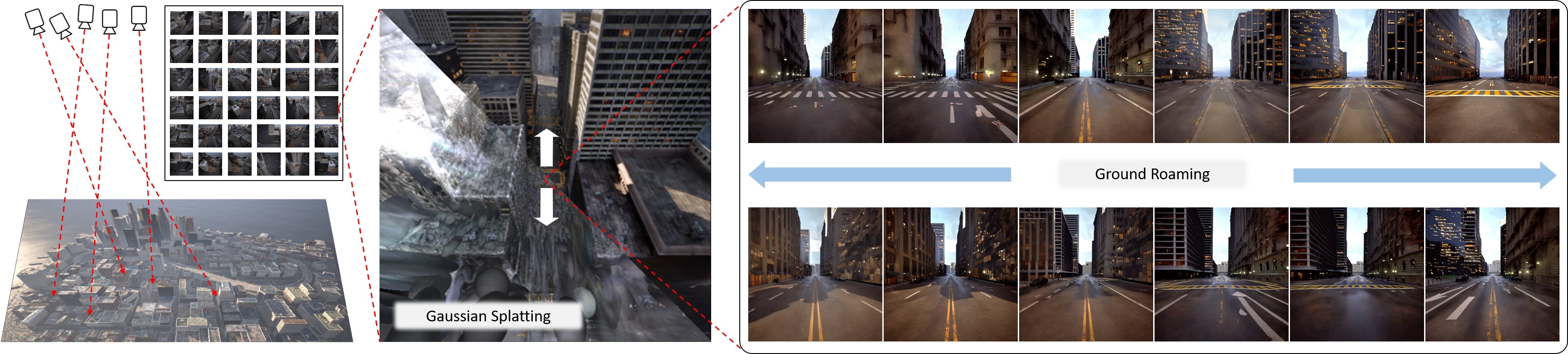}% ... an image
  \captionof{figure}{We proposed SkyEyes, a novel framework for efficient aerial-to-ground cross-view synthesis, transforming aerial imagery into realistic street view image sequence. This first-of-its-kind method for large-scale outdoor scenes combines 3D Gaussian Splatting with diffusion models to identify data gaps. Our constrained optimization strategy and View Consistent Module enable us to achieve images from entirely different perspectives compared to the input imagery, significantly enhancing the quality of ground-level view synthesis.}
  \vspace{0.5cm}
}
\makeatother

% Custom \maketitle definition
% \makeatletter
% \let\@oldmaketitle\@maketitle % Store original \maketitle
% \renewcommand{\@maketitle}{
%   % \onecolumn % Switch to single column mode
%   \@oldmaketitle % Call original \maketitle
%   \vspace{0.2cm}
%   \begin{center}
%     \includegraphics[width=0.95\linewidth]{fig/img/teaser.jpg} % Insert image
%     \captionsetup{type=figure} % Set caption type to figure
%     \caption{We proposed SkyEyes, a novel framework for efficient aerial-to-ground cross-view synthesis, transforming aerial imagery into realistic street view image sequence. This first-of-its-kind method for large-scale outdoor scenes combines 3D Gaussian Splatting with diffusion models to identify data gaps. Our constrained optimization strategy and View Consistent Module enable us to achieve images from entirely different perspectives compared to the input imagery, significantly enhancing the quality of ground-level view synthesis.}
%   \end{center}
%   % \vspace{0.5cm}
%   % \twocolumn % Switch to two-column mode
% }
% \makeatother

\maketitle

\begin{abstract}
    Integrating aerial imagery-based scene generation into applications like autonomous driving and gaming enhances realism in 3D environments, but challenges remain in creating detailed content for occluded areas and ensuring real-time, consistent rendering. In this paper, we introduce Skyeyes, a novel framework that can generate photorealistic sequences of ground view images using only aerial view inputs, thereby creating a ground roaming experience. More specifically, we combine a 3D representation with a view consistent generation model, which ensures coherence between generated images. This method allows for the creation of geometrically consistent ground view images, even with large view gaps. The images maintain improved spatial-temporal coherence and realism, enhancing scene comprehension and visualization from aerial perspectives. To the best of our knowledge, there are no publicly available datasets that contain pairwise geo-aligned aerial and ground view imagery. Therefore, we build a large, synthetic, and geo-aligned dataset using Unreal Engine. Both qualitative and quantitative analyses on this synthetic dataset display superior results compared to other leading synthesis approaches. See the project page for more results: \href{https://chaoren2357.github.io/website-skyeyes/}{chaoren2357.github.io/website-skyeyes/}.
\end{abstract}

\section{Introduction}
\label{sec:intro}

% Creating a high-quality, large-scale 3D simulation environment is crucial for applications like autonomous driving, gaming, and robotics. In these areas, photo-realism and diversity of the environments are of significant importance. In the conventional gaming industry pipeline, hand-crafting is still prevalent. Skilled artists need to put in intensive labor to create just one simulation environment. This process is not only costly but also often lacks realism in depicting the real-world distribution of landscapes, landmarks, and building types, underlining the necessity for methods that can effectively generate large-scale terrain and simulation environment based on real data. In this regards, aerial imagery becomes a key resource due to its broader coverage and ease of data acquisition. The process of transforming captured aerial views into detailed 3D terrain models, which can synthesize ground or street view sequences given a trajectory in the ground, known as aerial-to-ground cross-view synthesis, offers a promising path of solution.

Creating large-scale, high-quality 3D simulation environments is crucial for applications like autonomous driving, gaming, and robotics. However, traditional methods in the gaming industry often rely on labor-intensive handcrafting, which is both time-consuming and costly, limiting their scalability and realism in depicting real-world landscapes.

Aerial imagery plays a significant role in addressing this challenge due to its wide coverage and ease of acquisition. It provides a practical resource for generating large-scale 3D terrains and environments. However, transforming aerial views into accurate ground-level views remains a complex problem due to the significant differences between aerial and ground perspectives.

Existing techniques in related areas, while effective in some contexts, face significant limitations when applied to our task. First, methods like Structure from Motion (SfM)~\cite{schonberger2016structure}, Neural Radiance Fields (NeRF)~\cite{mildenhall2020nerf, xiangli2022bungeenerf, tancik2022block, xu2023gridguided, turki2022mega}, and 3D Gaussian Splatting 
  (3DGS)~\cite{kerbl20233d,guedon2023sugar,lu2024scaffold,ren2024octree,huang20242d} are designed for 3D reconstruction and novel view synthesis. These techniques work well when both the input and output belong to the same domain, such as generating novel ground-level views from multiple ground-level images. However, the aerial view captures the tops of buildings and large-scale layouts, revealing patterns and structures invisible from the ground, while the ground view focuses on building facades, entrances, and details like storefronts that are hidden from above. Since our task involves generating ground-level views from aerial images, which are in a different domain, these methods struggle to maintain high-quality outputs.

Second, satellite-to-ground inference techniques use satellite imagery to generate ground-level views~\cite{regmi2018cross, deng2018like, Lu_2020_CVPR, Li_2021_ICCV, toker2021coming, jang2021semantic, qian2023sat2density, li2024sat2scene, xu2024geospecific}. While these methods can maintain geometric consistency between the high-altitude satellite images and the inferred ground views, they are not required to capture the same level of geometric and textural detail that our task demands. The relatively high altitude of satellite images makes these techniques insufficient for generating precise and realistic ground-level views from lower-altitude aerial inputs.

Lastly, control-based image/video generation methods~\cite{ho2020denoising, saharia2022photorealistic, zhang2023adding, chen2023control,brooks2023instructpix2pix} use aerial views to guide the generation of corresponding ground-level images. While these approaches can generate ground views that align with individual aerial images, they often struggle to maintain geometric continuity across sequences. Even if they ensure consistency between a single aerial image and its corresponding ground view, they fail to preserve coherence when generating entire sequences of ground-level views from aerial image sequences.

\begin{figure*}
    \centering
    \includegraphics[width=0.99\linewidth]{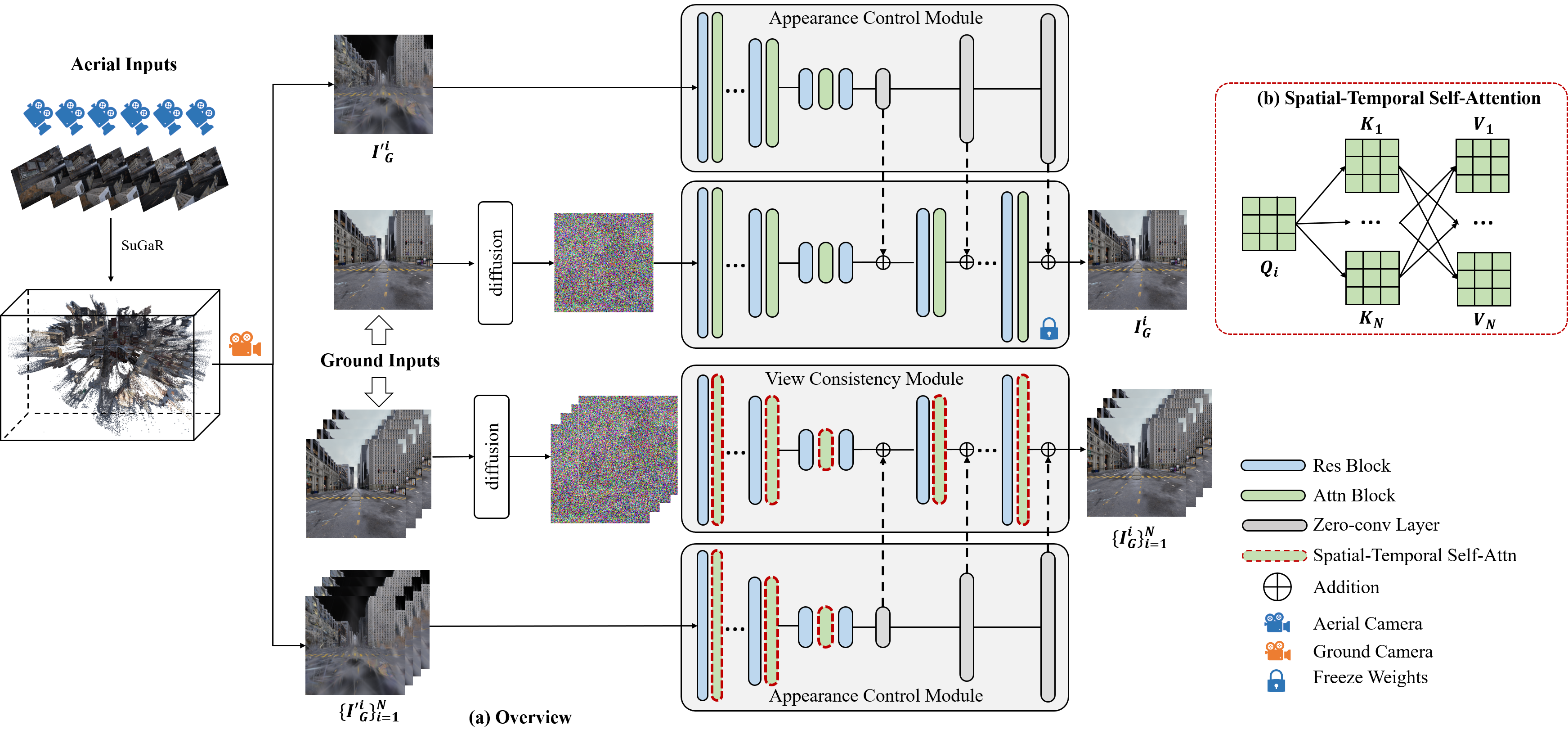}
    \caption{\textbf{(a) Overview of Skyeyes Pipeline:} Our approach commences with the utilization of SuGaR~\cite{guedon2023sugar}. This stage involves processing aerial images and camera poses to train the model for generating ground view priors. After that, we train an appearance control module to generate photo-realistic street images \textbf{(b) Spatial-Temporal Self-Attention Module:} In the final stage, our view consistency module integrates temporal modeling to ensure spatial and temporal coherence across different views. This module, akin to a spatial-temporal self-attention mechanism, guarantees the consistency and continuity of the scene's depiction across various perspectives. At inference time, given a sequence of ground view priors rendered from SuGaR~\cite{guedon2023sugar}, our view consistency module can generate photo-realistic and temporal consistent ground view sequence by denoising from pure Gaussian noise.}
    \label{fig:architecture}
\end{figure*}

To address the challenges outlined earlier, we introduce Skyeyes, a framework designed to generate photo-realistic and content-consistent ground-level image sequences from aerial image inputs. %\yajie{The writing of this paragraph should be polished and we can reduce the length} What sets SkyEyes apart are two key innovations that effectively address the identified problems. Firstly, SkyEyes is a seamless framework that transforms 2D aerial imagery into 3D terrain representation. This method not only maintains a 2D view space for ease of comprehension but also maximizes the preservation of known geometric and textural information derived from the aerial data. The dual approach ensures that the generated 3D models are both accurate and visually consistent with the original landscape. Secondly, unlike conventional full 3D methods, for example, 3D diffusion models, that require posed and densely collected data for training, SkyEyes leverages a 2D view-based method. This technique simplifies the training process and is more adaptable to varying scene scales. Moreover, it inherently maintains relative consistency between frames, providing a more reliable and cohesive modeling of terrain compared to methods that rely on extensive 3D data. These unique features of SkyEyes make it a more efficient and accurate tool for 3D terrain modeling, especially when dealing with large-scale and diverse landscapes.
% \Scott{Finish: these are steps rather than selling messages, so should be talked about after you sell why the overall idea is neat}
As depicted in Figure \ref{fig:architecture}, our approach first utilizes SuGaR, capitalizing superior detail retention ease of incremental updates, as well as its surface alignment nature compared with traditional 3DGS. This method effectively processes aerial view images and corresponding camera poses to train the model. Consequently, the optimized 3D Gaussians are then rendered from ground view perspectives, synthesizing ground view images that, while noisy, are imbued with a 3D-aware quality. Next, we implement an appearance control module designed to address the issue of preserving pixel accuracy in aerial views, a challenge noted in our integration of generative models. This module, functioning similarly to ControlNet~\cite{zhang2023adding} in the U-Net of the Stable Diffusion model, allows for controllable generation of photorealistic street view images. It effectively overcomes the limitations of pixel preservation in aerial imagery, ensuring a higher fidelity in the generated images. Finally, we introduce a view consistency module, which incorporates the concept of temporal modeling~\cite{ho2022video, chen2023control} into the appearance control module. This integration ensures that the content generated from each view in the ground sequence maintains spatial and temporal consistency. This approach directly addresses the challenge of maintaining a consistent view within a single scene, as highlighted in the discussion of generative models. By integrating these modules, we ensure that our terrain models not only capture the intricate details of the terrain but also maintain coherence and continuity across different views.

% Given a sequence of aerial images and their corresponding camera poses, we first train a 3DGS model and render the 3D Gaussians with a sequence of ground view cameras. For better rendering quality, we randomly select a ground view camera and remove the large ellipsoid by setting a customized threshold. \WT{Weird, rewrite the sentence.}. The rendered images are treated as a ground view appearance control of Stable Diffusion to generate photo-realistic ground view images. After that, a temporal consistency module is both added to the U-Net blocks of both ControlNet~\cite{zhang2023adding} and image diffusion model such that the generated ground view image sequence maintain its content coherence. The generated sequence is then sent back for finetuning of the initial 3DGS. 

% \Scott{Finish: if your results on real-data is good, you should mention that despite the data are synthetic, it well generalize in real scenes}

To the best of our knowledge, there are currently no publicly available datasets that provide pairwise geo-aligned aerial and ground level image sequences. However, such dataset is crucial for training our model. To tackle the problem of data scarcity, we extract large synthetic training data from open-source simulators including CARLA~\cite{Dosovitskiy17} and CitySample~\cite{Project} project developed in Unreal Engine 5. We customize sequence trajectories with respect to the location of different streets and render the scene with a spawnable camera. We will discuss more details of the dataset collection in Section~\ref{sec:exp-dataset}. We carried out extensive experiment on the extracted datasets to compare with the traditional methods and conduct ablation studies on different components of our proposed pipeline. Results and more details will be discussed in Section~\ref{sec:exp-result} and Section~\ref{sec:exp-ablation}. Both qualitative and quantitative analysis demonstrate that our method is superior than the other state-of-the-art frameworks. Code and both datasets will be released upon paper acceptance.

\section{Related Works}
\label{sec:related_work}

\subsection{Aerial-to-ground View Synthesis}
% Most of the previous research in this field predominantly utilizes the framework of generative adversarial networks (GANs)~\cite{goodfellow2014generative}  but differ in generating domain invariant features. Regmi et al.~\cite{regmi2018cross} and Deng et al.~\cite{deng2018like} proposed to learn the ground level RGB images based on conditional GANs, but they fail to take the geometric transformation into consideration thus lead to distorted outputs. \cite{Lu_2020_CVPR, Li_2021_ICCV} generate panoramic depth and semantic maps as condition of GANs by a geo-transformation layer. Toker et al.~\cite{toker2021coming} directly apply polar transformation to transform the satellite images into ground view. However, the output of these methods are mostly panoramic images, which has relative low resolution and preserves limited details. Some other methods like \cite{jang2021semantic} designed a semantic-attentive transformation module to align aerial features to the ground with respect to each semantic classes; however, the work mainly focus on rural area with very few semantic classes whereas urban areas have more complicated semantic classes. 
Previous research primarily employs GANs~\cite{goodfellow2014generative} for generating domain-invariant features. Regmi et al.~\cite{regmi2018cross} and Deng et al.~\cite{deng2018like} use conditional GANs to learn ground level RGB images but overlook geometric transformations, leading to distorted outputs. ~\cite{Lu_2020_CVPR, Li_2021_ICCV} generate panoramic depth and semantic maps using a geo-transformation layer. Toker et al.~\cite{toker2021coming} apply polar transformation for satellite to ground view conversion. However, these methods often produce panoramic images with low resolution and limited details. Jang et al.~\cite{jang2021semantic} design a semantic-attentive transformation module for aerial to ground alignment, but focus mainly on rural areas with fewer semantic classes, while urban areas present more complex challenges. Although methods like~\cite{Lu_2020_CVPR, Li_2021_ICCV} produce photo-realistic street view images, they refrain from the controllable generation of textures guided by aerial view priors. Therefore, their problem settings are different from ours. To incorporate more prior knowledge of satellite imagery, Xu et al.~\cite{xu2024geospecific} incorporate both texture and high-frequency layout as condition of the ground view panorama generation model, but it fails to address the time consistency problem across different frames.
\subsection{Large Scale Novel View Synthesis}

Novel view synthesis, primarily driven by Neural Radiance Fields (NeRF) ~\cite{mildenhall2020nerf}, has seen significant advancements through deep learning, enabling diverse scene representations and new view rendering.  However, traditional NeRF struggles with large-scale environments due to intense memory and computational demands and challenges in handling transient objects. To address these limitations, recent developments have focused on adapting NeRF techniques for ground-level and aerial-level perspectives. From ground-level perspective, Block-NeRF ~\cite{tancik2022block} subdivides large environments into smaller, independently trained NeRFs, but it still encounters issues like temporal inconsistencies and less detailed reconstructions of distance objects.  Scalable Urban Dynamic Scenes(SUDS) ~\cite{turki2023suds} offers a novel approach by factorizing scenes into static, dynamic, and far-field components using separate hash tables to solve challenges in dynamic elements. But the work  only focuses on urban settings.  StreetSurf ~\cite{guo2023streetsurf}, on the other hand, separates unbounded spaces into multi-view segments and utilizes hyper-cuboid hash-grids and a road surface initialization scheme to enhance representation.  However, this method is primarily tailored for autonomous driving datasets and may under-perform in poor lighting conditions.  Street Gaussians ~\cite{yan2023streetgaussians} offers a different approach for urban scenes using 3D Gaussians, enabling swift object and background composition, but are limited to grid dynamics.  Despite all these innovations, handling transient objects still remains as a challenge in the field.

%%%%%%%%%%%%%%%%%%%%%%%%%%%%%%%%%%%%%%%%%%%%%%%%%%%%%%%%%%%%
%Several studies have successfully scaled Neural Radiance Fields (NeRF)-based methodologies to accommodate larger scenes, particularly those captured from aerial perspectives. %
In the aerial perspective domain, several studies have extended NeRF-based methodology to encompass larger scenes.
% Xiangli et al.~\cite{xiangli2022bungeenerf} trained the NeRF model using a progressive training strategy that allowed the model to grow by appending new blocks to model multi-scale scenes with varying views. Xu et al.~\cite{xu2023gridguided} proposed a two-branch architecture that uses a feature grid representation to capture the scene geometry and appearance, guiding the NeRF for rendering with efficiency for large city scenes. Turki et al.~\cite{turki2022mega} introduce a geometric clustering approach to decompose the scene into NeRF submodules that could be trained parallelly. However, these NeRF-based methods encounter difficulties in producing realistic images when the inference involves viewpoints that significantly differ from those of the input images. This limitation arises due to the constraint imposed by the available range of input perspectives.
Xiangli et al.~\cite{xiangli2022bungeenerf} employed a progressive training strategy for NeRF models to handle multi-scale scenes. Xu et al.~\cite{xu2023gridguided} developed a two-branch architecture with a feature grid for efficient rendering in large city scenes. Turki et al.~\cite{turki2022mega} proposed a geometric clustering method for parallel training of NeRF submodules. However, these NeRF-based approaches struggle with realistic image generation from significantly different viewpoints due to the limited range of input perspectives.

% To enhance the generalization capabilities of Neural Radiance Fields (NeRF), significant research has been directed towards developing methodologies for sparse or even single-view NeRF applications. Yu et al.~\cite{yu2021pixelnerf} and Lin et al.~\cite{lin2022vision}. tried to address this issue by introducing NeRF-based learning frameworks that conditioned on feature maps that encoded the novel view to be generated. Wu et al.~\cite{wu2023reconfusion} approached this problem by leveraging a diffusion prior which guided a NeRF-based pipeline for novel view synthesis. Despite these approaches showing incredible results, they tend to induce blurriness when generating ground-view images with aerial-view images.

\subsection{Controllable Image and Video Diffusion Model}

Diffusion model and latent diffusion model has exhibited their effectiveness in conditional image generation. By simply adding a text prompt, methods like Imagen~\cite{saharia2022photorealistic} and Stable Diffusion (SD) can achieve the ideal customization of content synthesis. The controllable image generation has been largely extended with the advent of ControlNet~\cite{zhang2023adding}, which allows additional condition to SD models such as depth, pose and segmentation maps. Established on ControlNet, Control-A-Video (CAV)~\cite{chen2023control} generates both controllable and content-consistent video based on sequence of control maps and text conditions. Apart from the traditional 3D U-Net proposed by video diffusion model~\cite{ho2020denoising}, one of the main contributions of CAV is the introduction of motion-adaptive noise initializer, which preserves the latent similarity between frames as appose to the random Gaussian noise.

\section{Skyeyes}

In this section, we elaborate on the details of Skyeyes. Given a sequence of aerial imagery $\{I_A^i\}_{i=1}^N$ and corresponding camera pose $\{W_A^i\}_{i=1}^N$, our goal is to synthesize a sequence of ground image $\{I_G^i\}_{i=1}^N$ conditioned on ground camera poses $\{W_G^i\}_{i=1}^N$, where $N$ is the number of selected frames in a sequence. The synthesized images $\{I_G^i\}_{i=1}^N$ should maintain its content coherence.

The overall architecture of Skyeyes is shown in Figure~\ref{fig:architecture}. We will first introduce the preliminaries of our proposed method in Section~\ref{sec:preliminary}, which includes 3D Gaussian Splatting and latent diffusion model/ControlNet. Then we will introduce our method in two steps. The first step, which involves the Appearance Control Module, is presented in Section ~\ref{sec:appearance_control_module}. The second step, concerning the View Consistency Module, is detailed in Section~\ref{sec:view_consistency_module}.

\subsection{Preliminary}
\label{sec:preliminary}
\subsubsection{Surface-Aligned 3D Gaussian Splatting(SuGaR)}
\label{sec:sugar}
3DGS~\cite{kerbl20233d} models a scene as a set of differentiable 3D Gaussians that could be easily rendered with tile-based rasterization.  Each Gaussian is parameterized by a center point $\mu_g$ and a covariance matrix $\Sigma_g$:

\begin{equation}
\label{eqn:gaussian-mu}
    G(x)=e^{-\frac{1}{2}(x-\mu_g)^T\Sigma_g^{-1}(x-\mu_g)}
\end{equation}

% The center point $\mu_g \in \mathbb{R}^3$ is usually initialized with the sparse points obtained from Structure-from-motion (SFM). The covariance matrix $\Sigma$ is formulated by a rotation matrix $R$ and scaling matrix $S$ to maintain its positive semi-definite:
% \begin{equation}
% \label{eqn:gaussian-sigma}
%     \Sigma = RSS^TR^T
% \end{equation}

% All of the parameters, including $R$, $S$, $\mu$ and Spherical Harmonics parameter $c\in\mathbb{R}^k$, opacity $\alpha\in\mathbb{R}$ are optimized by differentiable rendering. Compared to ray marching of NeRF~\cite{mildenhall2020nerf}, 3DGS~\cite{kerbl20233d} introduces the splatting of Gaussians to the image plane. Given a camera pose matrix $W$, the covariance matrix in 2D image plane can be formulated as:
% \begin{equation}
% \label{eqn:gaussian-sigma2d}
%     \Sigma^{2D}=JW\Sigma W^TJ^T
% \end{equation}

% \noindent where $J$ is the Jacobian of projective transformation matrix. 
% and~\ref{eqn:gaussian-sigma2d}
When rendering, the color and opacity of all the Gaussians are calculated by Equation~\ref{eqn:gaussian-mu}. The final pixel color $C$ is computed by blending all the 2D Gaussians that contributes to the pixel:
\vspace{-15pt}
\begin{equation}
    \label{eqn:gaussian-rendering}
    C=\sum_{i\in N}c_i\alpha_i\prod_{j=1}^{i-1}(1-\alpha_j)
    \vspace{-5pt}
\end{equation}
\noindent where $c_i$ and $\alpha_i$ are the view dependent color and opacity of the Gaussian. For more details, we recommend the original work from~\cite{kerbl20233d}.

In the SuGaR framework that utilizes 3D Gaussian Splatting, the process begins by incorporating a loss term based on the signed-distance field (SDF). This loss term, represented by Equation \ref{eqn:sdf-loss}, ensures the alignment of 3D Gaussians with the scene's surface during optimization. A rough mesh is extracted from the aligned Gaussians, and both the mesh and the 3D Gaussians situated on the mesh surface are optimized jointly using Gaussian Splatting rendering, resulting in a new set of Gaussians that are tied to an editable mesh. 
\vspace{-5pt}
\small % or \scriptsize for an even smaller font size
\begin{equation}
    \label{eqn:sdf-loss}
    R = \frac{1}{|P|} \sum_{p \in P} \left| \frac{<p-\mu_{g^*},n_{g^*}>}{s_{g^*}} - \frac{<p-\mu_g,n_g>}{s_g} \right|
    \vspace{-5pt}
\end{equation}
\normalsize % Reset to normal size after the equation

Here, \( R \) denotes the residual error across a set of sample points \( P \). \( s_g \) is the smallest scaling factor of Gaussian \( g \), which signifies how flat the Gaussian is—approaching zero implies increased flatness. The parameters \( \mu_{g^*} \) and \( s_{g^*} \) are the optimal Gaussian parameters that align best with the scene's surface.

Moreover, the methodology seeks to avoid semi-transparent Gaussians to accurately describe the scene's surface, hence, the opacity coefficient \( \alpha_g \) is set to 1 for any Gaussian \( g \). More details can be obtained in original paper ~\cite{guedon2023sugar}.

\subsubsection{Latent Diffusion Models and ControlNet}
Compared to diffusion models~\cite{ho2020denoising,song2020score}, latent diffusion models~\cite{rombach2022high} synthesize features of images in a latent space defined by a pre-trained autoencoder. A common schema is to add textual inputs into image generation by converting a text prompt into embeddings $c_{text}$. This is usually achieved by a CLIP-based transformer for text encoding. Given an Image $I$ and encoder $\mathcal{E}$, the initial latent feature $z_0=\mathcal{E}(I)$ is perturbed by  a sequence of Gaussian noise such that after $T$ steps  the latent feature $z_T$ fall in a standard Gaussian distribution $\mathcal{N}(0, 1)$. The objective of latent diffusion model is to optimize a denoising process formulated by a U-Net architecture:
\begin{equation}
    \mathcal{L}_{LDM}=\mathbb{E}_{z_0,c_{text},t,\epsilon\sim\mathcal{N}(0,1)}\Big[\|\epsilon-\epsilon_\theta(z_t,c_{text},t)\|_2^2\Big]
\end{equation}
\noindent Here $t=1...T$ is the time embedding during denoising process. $\epsilon$ is a standard normal distribution and $\epsilon_\theta$ is the neural network parameterized by $\theta$. 

ControlNet~\cite{zhang2023adding} further boost the controlability of latent diffusion model by adding a specific image condition such as depth or semantic map. The downsampling blocks and middle block of ControlNet is a trainable copy of Stable Diffusion~\cite{rombach2022high} whereas its main contribution is to add a series of zero-convolutions whose outputs are added to the skipped connection of Stable Diffusion U-Nets. Suppose the task-specific image condition is denoted as $c_f$, the objective is formulated as follows:
\vspace{-5pt}

\scriptsize % or \footnotesize
\begin{equation}
    \mathcal{L}_{Control}=\mathbb{E}_{z_0,c_{text},c_f,t,\epsilon\sim\mathcal{N}(0,1)}\Big[\|\epsilon-\epsilon_\theta(z_t,c_{text},c_f,t)\|_2^2\Big]
    \vspace{-5pt}
\end{equation}
\normalsize % Reset to normal size after the equation

\begin{figure*}
    \centering
    \begin{subfigure}{0.215\textwidth}
        \centering
        \includegraphics[width=\linewidth]{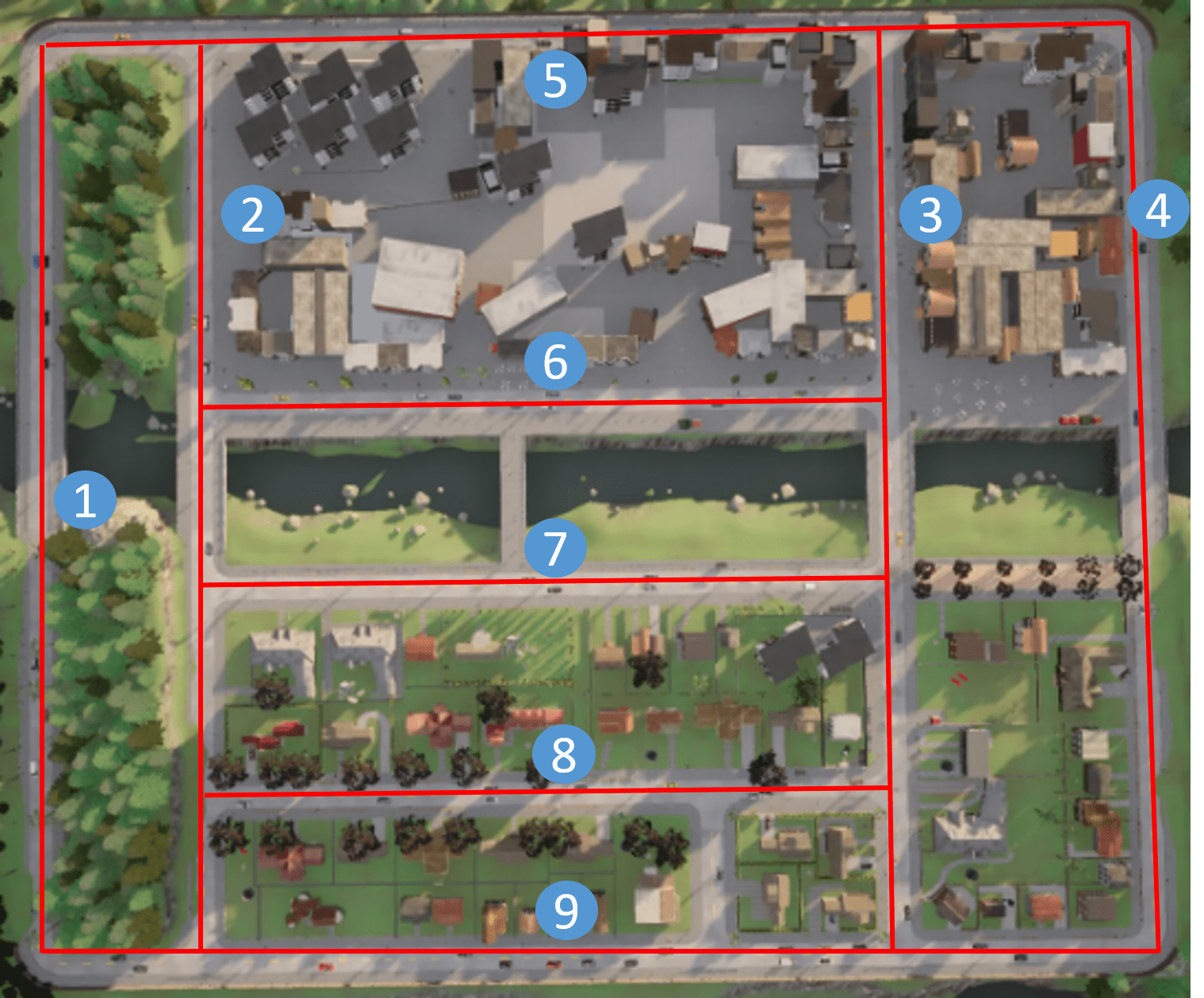}
        \caption{CARLA Town 01}
        \label{fig:carla-map}
    \end{subfigure}
    \begin{subfigure}{0.373\textwidth}
        \centering
        \includegraphics[width=\linewidth]{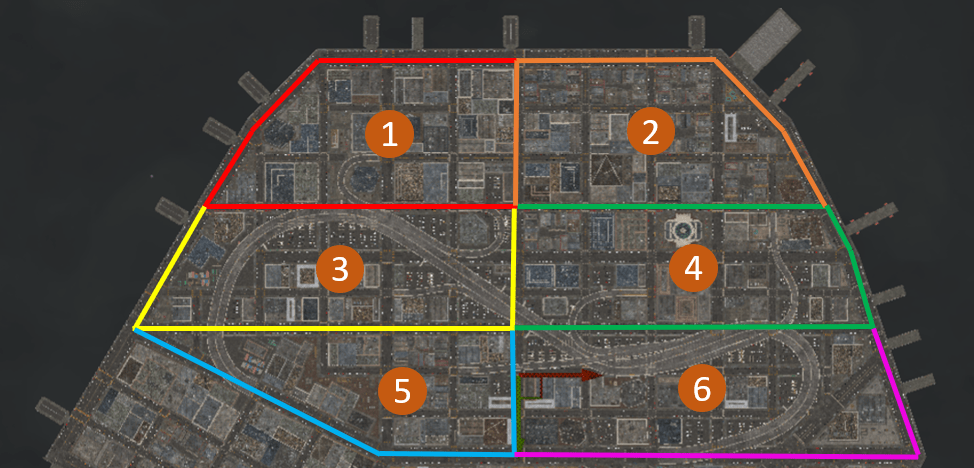}
        \caption{Region split of City Sample}
        \label{fig:cs-region}
    \end{subfigure}
    \begin{subfigure}{0.345\textwidth}
        \centering
        \includegraphics[width=\linewidth]{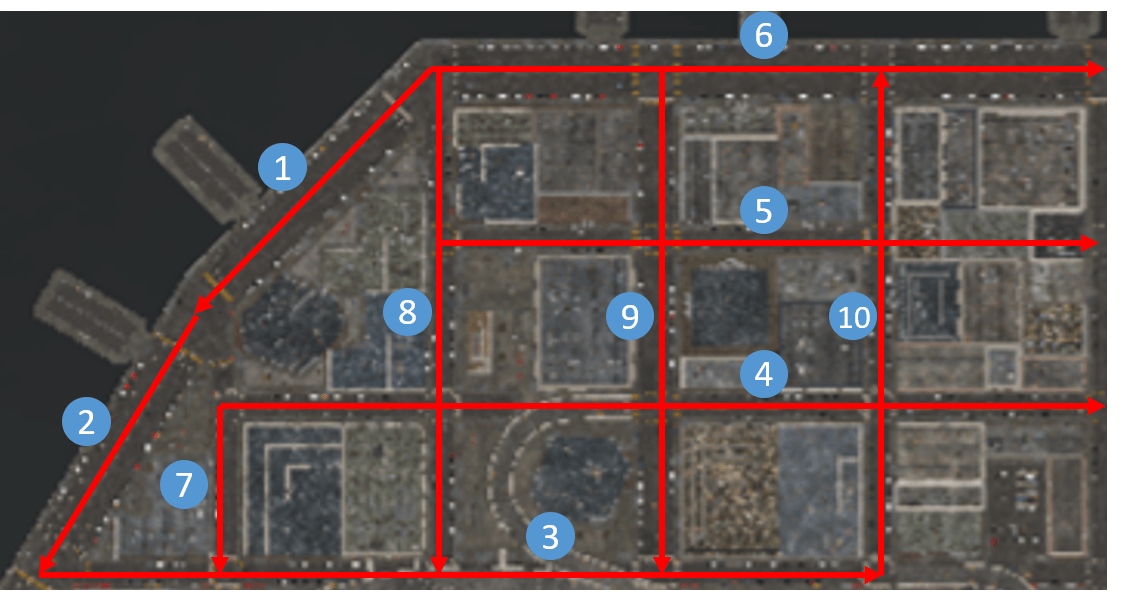}
        \caption{City Sample Region 1}
        \label{fig:cs-region1-lane}
    \end{subfigure}
    \label{fig:dataset-maps}
    \caption{Comprehensive visual representation of the data extraction process from CARLA and City Sample Project. 
    %These figures collectively illustrate the strategic selection of lanes and regions within urban simulation environments. Each subfigure showcases the methodical approach used in delineating specific areas for data collection, demonstrating the meticulous planning involved in acquiring diverse and relevant datasets for our study.
    }
    \vspace{-15pt} % Reduce the vertical space

\end{figure*}
\subsection{Appearance Control Module}
\label{sec:appearance_control_module}
The objective of appearance control module is to leverage the controllable image generation ability of LDM~\cite{rombach2022high} to generate $\{I_G^i\}_{i=1}^N$ conditioned on $\{I_A^i\}_{i=1}^N$. This naturally leads us to the architecture of ControlNet~\cite{zhang2023adding}. However, ControlNet incorporates task-specific images (depth, semantic, etc.) as conditions which are not accessible for the generation of ground view images. One straightforward way is to project pixels of $I_A^i$ from image space to world space and project back to the image space of $\{I_G^i\}$, but this requires accurate geometry of $I_A^i$, which is usually hard to obtain in real application and a coarse depth estimation will exhibit large view distortion and appearance distinction.

In appearance control module, we propose to leverage SuGaR~\cite{guedon2023sugar} to construct the scene given both $\{I_A^i\}_{i=1}^N$ and $\{W_A^i\}_{i=1}^N$. Compared with traditional 3DGS~\cite{kerbl20233d} tends to align the 3D Gaussians with the surface of the object. The optimized 3D Gaussians are rendered with ground view cameras $\{W_G^i\}_{i=1}^N$ to synthesize ground view control maps $\{I_G^{\prime i}\}$:
\vspace{-10pt}
\begin{equation}
    I_G^{\prime i}=\mathcal{R}(G(\{I_A\}_{i=1}^N,\{W_A\}_{i=1}^N), W_G^i)
    \vspace{-5pt}
\end{equation}

\noindent where $\mathcal{R}$ is the Gaussian splatting renders. In this step, we first initialize the point cloud using Structure-from-Motion (SfM). The inputs for SfM are the sets $\{I_A^i\}_{i=1}^N$ and $\{W_A^i\}_{i=1}^N$, as accurate camera poses are essential for geo-registration. 

\subsection{View Consistency Module}
\label{sec:view_consistency_module}
With the appearance control module discussed in Section~\ref{sec:appearance_control_module} we are able to synthesize a photo-realistic ground view image from a noisy 3D Gaussian prior. Nevertheless, how to ensure content consistency across all the views in a sequence remains a challenging problem. The general purpose of appearance control module is to refine the blurry regions and in-paint the unseen regions with the powerful content generation ability of latent diffusion model. However, how the regions are refined and in-painted may have thousands of explanations. In this work, we are inspired by vid2vid-Zero~\cite{wang2023zero} and Control-A-Video~\cite{chen2023control} to propose a view consistency module (VCM) for a sequence of generated ground views. VCM is integrated to the up-sampling and down-sampling blocks of  LDM to maintain both spatial and temporal consistency. Illustrated by Figure~\ref{fig:architecture}, suppose the self-attention calculated by the pre-trained LDM is given by:
\vspace{-10pt}
\begin{equation}
    Self\_Attn(\cdot)=softmax\Big(\frac{QK^T}{\sqrt{d}}\Big)\cdot V
    \vspace{-5pt}
\end{equation}
\noindent where $Q,K,V$ are the query, key, value features of the spatial features $x$ such that $Q=W^Qx,K=W^Kx,V=W^Vx$, and $W^Q,W^K,W^V$ are the corresponding learnable projection matrix. Instead of simply considering the spatial self-attention across the feature maps, we incorporate the spatial-temporal self-attention across the frames, where the projection matrices are shared for all the frames:
\vspace{-5pt}
\begin{equation}
    Q=W^Qx_i, K=W^Kx_{1:F}, V=W^Vx_{1:F}
    \vspace{-5pt}
\end{equation}
\noindent where $x_i$ is the query frame, and $x_{1:F}$ are the concatenation of all the frame in a sequence, i.e. $x_{1:F}=[x_1,x_2,...,x_F]$. By attending both the spatial features across the feature map and the temporal features across all the frames, we find it effective to alleviate the discrepancy between each sampling process of LDM and ControlNet, with more details are discussed in Section~\ref{sec:ablation-view-consistency-module}.

With memory efficiency as well as long sequence generation purpose, we condition the generation process conditioned on the latent space of the first frame. Specifically, we add noise to each frame of the random sampled sequence except for the first frame. By this training scheme, the diffusion model can learn to generate the subsequent frames based on the first frame. The objective is formulated as follows:

\vspace{-5pt}
\scriptsize
\begin{equation}
    \mathcal{L}_{VCM}=\mathbb{E}_{z_0,c_{text},c_f,t,\epsilon\sim\mathcal{N}(0,1)}\Big[\|\epsilon-\epsilon_\theta(z_t,c_{text},c_f,t,z^1)\|_2^2\Big]
    \vspace{-5pt}
\end{equation}
\normalsize
\noindent where $z^1$ is the latent feature of the first frame. At inference time, we first generate the first frame $x^1$ with our appearance control module and use $x^1$ as the condition to autoregressively generate the subsequence frames:
\vspace{-3pt}
\begin{equation}
    x_{2:F}=VCM(z_{2:F},c_{text}, c_f, \mathcal{E}(x^1))
    \vspace{-3pt}
\end{equation}

This proposed method will refrain the diffusion model from memorizing all the frames in a video thus achieves both memory efficiency and long sequence generation targets. During training, we randomly sample a consecutive of 12 frames within a long ground view sequence and train the diffusion U-Net with the first-frame conditioning method.

\begin{figure*}
    \centering
    \begin{subfigure}{0.9\linewidth}
        \centering
        \includegraphics[width=\linewidth]{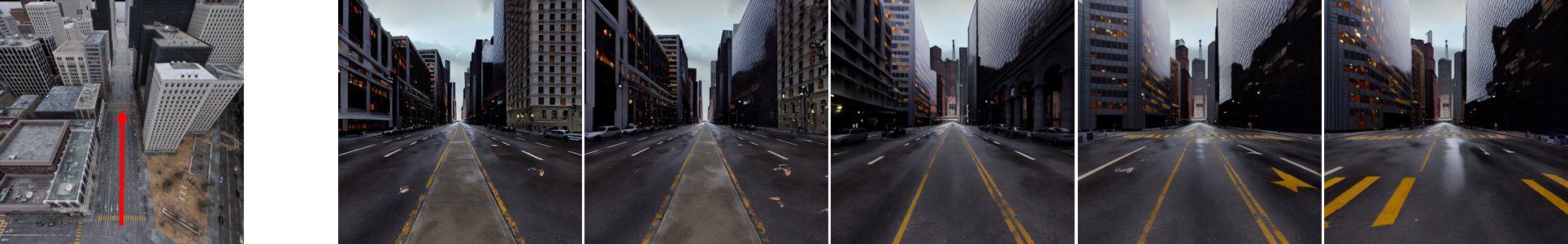}
    \end{subfigure}
    % \vspace{0.3cm}
    \begin{subfigure}{0.9\linewidth}
        \centering
        \includegraphics[width=\linewidth]{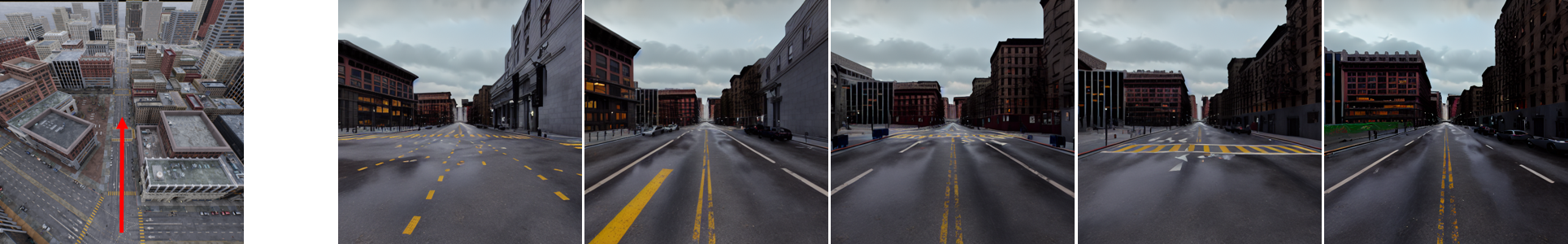}
    \end{subfigure}
    % \vspace{0.3cm}
    \begin{subfigure}{0.9\linewidth}
        \centering
        \includegraphics[width=\linewidth]{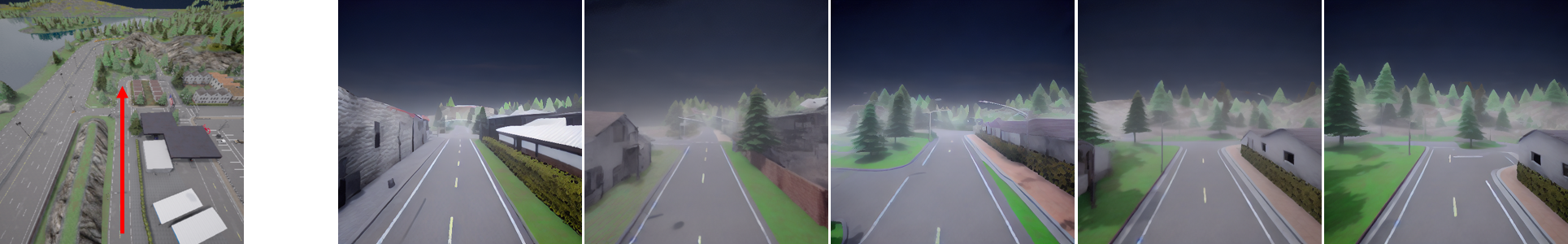}
    \end{subfigure}
    % \vspace{0.3cm}
    \begin{subfigure}{0.9\linewidth}
        \centering
        \includegraphics[width=\linewidth]{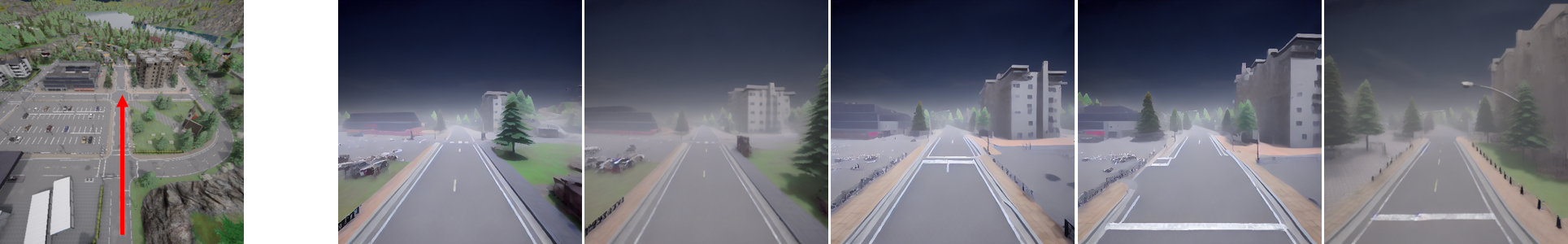}
    \end{subfigure}
    \caption{\textbf{Qualitative Results.} Conditioned on aerial images (leftmost column), our method synthesizes realistic and view-consistent ground view sequences. The first two rows are from the CitySample dataset, and the last two from the CARLA dataset. We strongly recommend checking the supplementary material for more results.}
    \label{fig:qual-fig1}
    \vspace{-0.5cm} 

\end{figure*}

\section{Experiments}
\label{sec:exp}
\subsection{Dataset Collection}
\label{sec:exp-dataset}
We utilized two distinct scene simulation platforms, CARLA Simulator~\cite{Dosovitskiy17} and CitySample~\cite{Project} from Unreal Engine 5, to generate geo-aligned aerial and street view datasets for detailed and complex urban or rural environments.

\paragraph{CARLA Simulator} is an open-source platform designed for the development, training, and validation of autonomous driving systems. The sequences are extracted from varies maps including Town01, Town02, Town03, Town04 and Town05 where we manually locate the start and end point of each lane (See Fig.~\ref{fig:carla-map} for an example of lane selection of Town 01). %Within each lane, we spawn cameras to capture a color image for every 2 meters. Camera positioning is automatically optimized by CARLA to adhere to constraints like maintaining a safe distance from buildings, and the camera orientation is adjusted to face the direction of travel. For each point sampled on the lane, we set the yaw value of camera rotation to vary within $k\pi/4$, where $k=0...7$. The altitude of aerial sequence is set to be 52 meters while the altitude of ground sequence is 2 meters. Pitch value of camera rotation is set to be -45 degrees for aerial views whereas 0 for ground views. For training-evaluation purposes, we set all the extracted data from Town04 for evaluation and all the rest for training. 

\paragraph{CitySample} project is created by Ubisoft in Unreal Engine 5. We selected the smaller city level within this project for data extraction. Given the expansive scale of the map, we segmented it into multiple regions (refer to \ref{fig:cs-region} for details) and appoint multiple lanes inside a region (refer to~\ref{fig:cs-region1-lane} for details). 
% We manually determine the start and end point of each lane. Within each lane, we spawn cameras to capture various signals such as color image, depth and semantic maps for every 2 meters. For each point sampled on the lane, we set the yaw value of camera rotation to vary within $k\pi/4$, where $k=0...7$. The altitude of aerial sequence is set to be 100 meters while the altitude of ground sequence is 2 meters. Pitch value of camera rotation is set to be -45 degrees for aerial views whereas 0 for ground views. 
%This data extraction protocol mirrors the data extraction strategy employed in the CARLA Simulator, where the starting and ending points of each lane were manually determined (as exemplified in Figure~\ref{fig:cs-region1-lane}). The configuration of camera poses in this environment closely aligns with those in CARLA, with the notable distinction that aerial sequences are captured at an altitude of 100 meters. We choose region 5 as the test set, and region 1, 2, 3, 4 and 6 as the training set.
%To enhance the efficiency and quality of data collection, we incorporated a plugin developed by MatrixCity~\cite{li2023matrixcity}, which facilitated the capture of color images and depth maps. Consistent with our training-evaluation objectives, we designated all data extracted from region 4 as our evaluation set, while data from other regions were allocated for training purposes. 

Please refer to our supplementary materials for a more detailed description of data collection process.

\subsection{Implementation Details}
We organized aerial and ground view images by distinct lanes, treating each lane's set as a sequential dataset with ground-level images as primary input. We first train SuGaR on an NVIDIA RTX 4090 GPU for 15K iterations, resizing images to $512 \times 512$. Then, priors $\{I^{\prime i}_G\}$ are rendered from ground view poses for all sequences. These priors are used to train the appearance control module for 30K iterations, with a batch size of 32 on 4 NVIDIA A100 GPUs. We condition the diffusion model with a uniform text prompt, "\texttt{a realistic street view image}". Finally, we freeze the appearance control module and train the view consistency module for 3K iterations with a batch size of 2, sampling 12 frames from a large sequence for generalization.

\subsection{Baselines and Metrics}
\paragraph{Baselines} 
%To tackle the task of producing high-quality terrain effectively, it's crucial to investigate innovative methods. Established techniques such as Structure from Motion, NeRF, and 3DGS offer promising options. Consequently, for our foundational analysis, 
% We have chose MVS~\cite{schoenberger2016sfm}, Instant NGP~\cite{mueller2022instant}, the original 3DGS, and SuGaR~\cite{guedon2023sugar}. While other methods, such as those cited in ~\cite{Lu_2020_CVPR, Li_2021_ICCV}, focus on satellite-to-ground synthesis, these approaches primarily extract limited geometric information from overhead satellite imagery. In contrast, our task demands not only finer geometric details but also consistent and high-resolution texture generation. The requirements for our dataset and objectives are significantly more stringent, making the selected baselines more appropriate for the comprehensive evaluation of high-quality terrain generation.

As mentioned in Section~\ref{sec:intro}, we chose three types of baseline methods that aligned with our task
For 3D reconstruction methods,  we chose MVS~\cite{schonberger2016structure}, NeRF~\cite{mildenhall2020nerf}, 3DGS~\cite{kerbl20233d} and SuGaR~\cite{guedon2023sugar}; Geospecific View Generation (GVG)~\cite{xu2024geospecific} for satellite-to-ground related baseline; ControlNet~\cite{zhang2023adding} and InstructPix2Pix~\cite{brooks2023instructpix2pix} for control-based image/video generation baseline.

\begin{figure*}
    \centering

    % % First row
    % \begin{minipage}[c]{.03\linewidth}
    %     \rotatebox[origin=c]{90}{MVS~\cite{schoenberger2016sfm}}
    % \end{minipage}%
    % \hfill
    % \begin{minipage}[c]{.42\linewidth}
    %     \begin{subfigure}[b]{\linewidth}
    %         \includegraphics[width=\linewidth]{fig/img/experiments/qual-fig2/seq_0009_mvs_full.png}
    %     \end{subfigure}
    % \end{minipage}%
    % \hfill
    % \begin{minipage}[c]{.42\linewidth}
    %     \begin{subfigure}[b]{\linewidth}
    %         \includegraphics[width=\linewidth]{fig/img/experiments/qual-fig2/seq_0156_mvs_full.png}
    %     \end{subfigure}
    % \end{minipage}

    % % Second row
    % \begin{minipage}[c]{.03\linewidth}
    %     \rotatebox[origin=c]{90}{NeRF~\cite{mueller2022instant}}
    % \end{minipage}%
    % \hfill
    % \begin{minipage}[c]{.42\linewidth}
    %     \begin{subfigure}[b]{\linewidth}
    %         \includegraphics[width=\linewidth]{fig/img/experiments/qual-fig2/seq_0009_nerf_full.png}
    %     \end{subfigure}
    % \end{minipage}%
    % \hfill
    % \begin{minipage}[c]{.42\linewidth}
    %     \begin{subfigure}[b]{\linewidth}
    %         \includegraphics[width=\linewidth]{fig/img/experiments/qual-fig2/seq_0156_nerf_full.png}
    %     \end{subfigure}
    % \end{minipage}

    % Third row
    \begin{minipage}[c]{.02\linewidth}
        \rotatebox[origin=c]{90}{SuGaR}
    \end{minipage}%
    \hfill
    \begin{minipage}[c]{.48\linewidth}
        \begin{subfigure}[b]{\linewidth}
            \includegraphics[width=\linewidth]{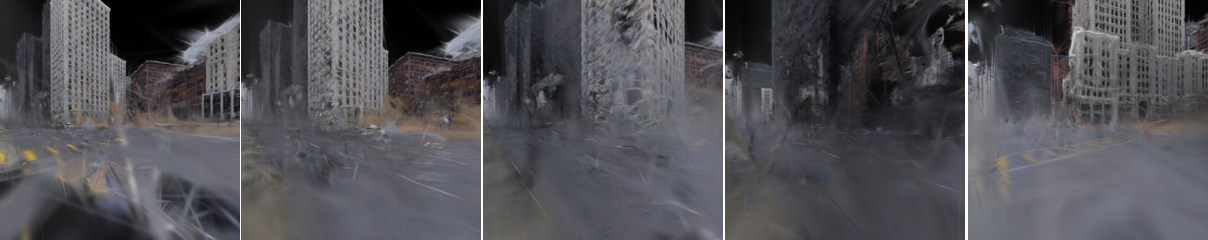}
        \end{subfigure}
    \end{minipage}%
    \hfill
    \begin{minipage}[c]{.48\linewidth}
        \begin{subfigure}[b]{\linewidth}
            \includegraphics[width=\linewidth]{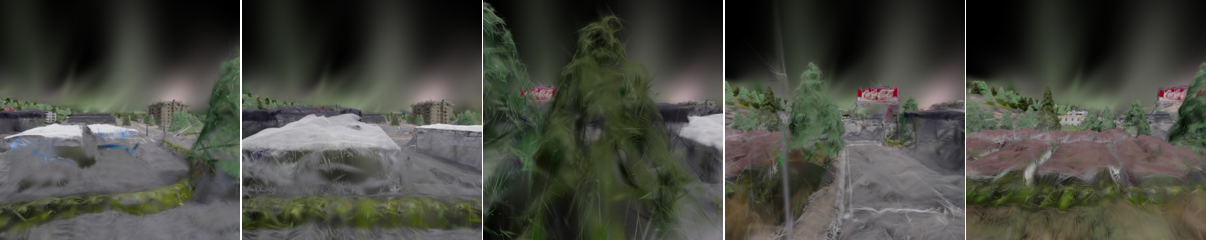}
        \end{subfigure}
    \end{minipage}

    % another row
    \begin{minipage}[c]{.02\linewidth}
        \rotatebox[origin=c]{90}{ControlNet}
    \end{minipage}%
    \hfill
    \begin{minipage}[c]{.48\linewidth}
        \begin{subfigure}[b]{\linewidth}
            \includegraphics[width=\linewidth]{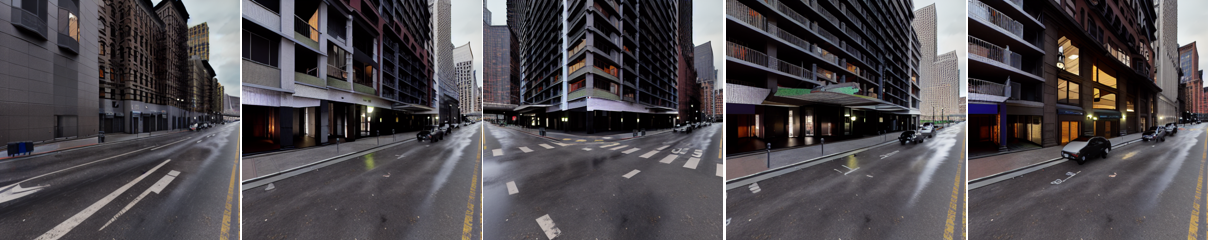}
        \end{subfigure}
    \end{minipage}%
    \hfill
    \begin{minipage}[c]{.48\linewidth}
        \begin{subfigure}[b]{\linewidth}
            \includegraphics[width=\linewidth]{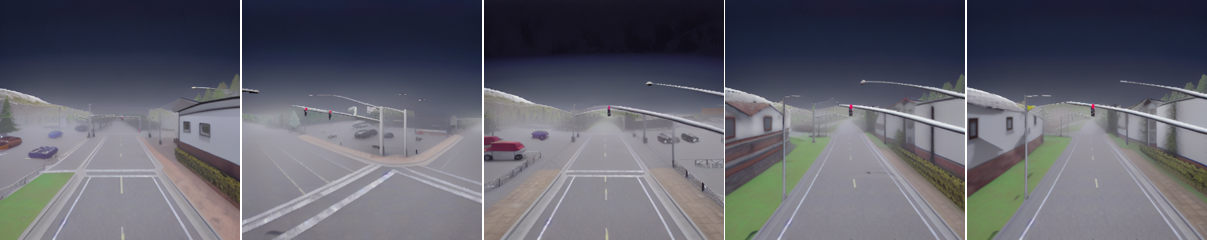}
        \end{subfigure}
    \end{minipage}
    
    % another row
    \begin{minipage}[c]{.02\linewidth}
        \rotatebox[origin=c]{90}{GVG}
    \end{minipage}%
    \hfill
    \begin{minipage}[c]{.48\linewidth}
        \begin{subfigure}[b]{\linewidth}
            \includegraphics[width=\linewidth]{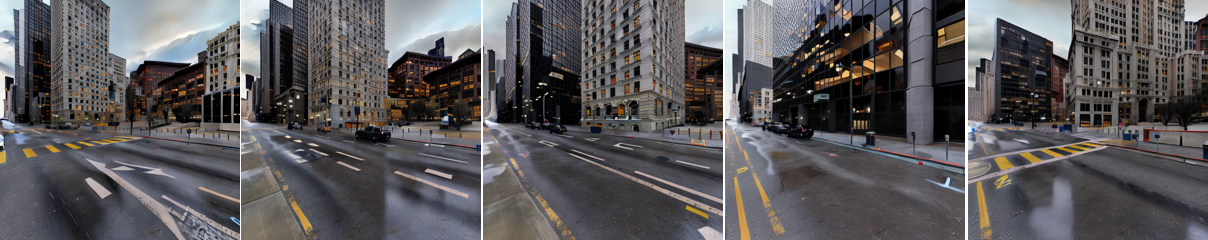}
        \end{subfigure}
    \end{minipage}%
    \hfill
    \begin{minipage}[c]{.48\linewidth}
        \begin{subfigure}[b]{\linewidth}
            \includegraphics[width=\linewidth]{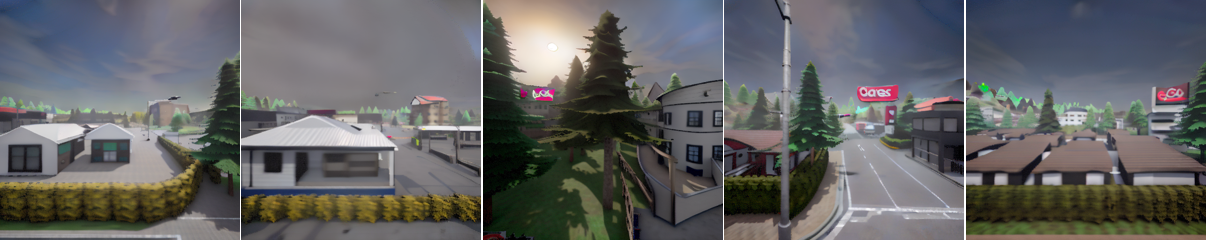}
        \end{subfigure}
    \end{minipage}

    % Fourth row
    \begin{minipage}[c]{.02\linewidth}
        \rotatebox[origin=c]{90}{Ours}
    \end{minipage}%
    \hfill
    \begin{minipage}[c]{.48\linewidth}
        \begin{subfigure}[b]{\linewidth}
            \includegraphics[width=\linewidth]{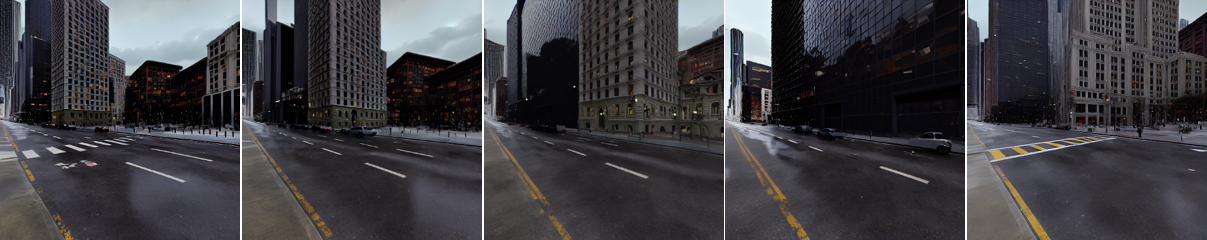}
        \end{subfigure}
    \end{minipage}%
    \hfill
    \begin{minipage}[c]{.48\linewidth}
        \begin{subfigure}[b]{\linewidth}
            \includegraphics[width=\linewidth]{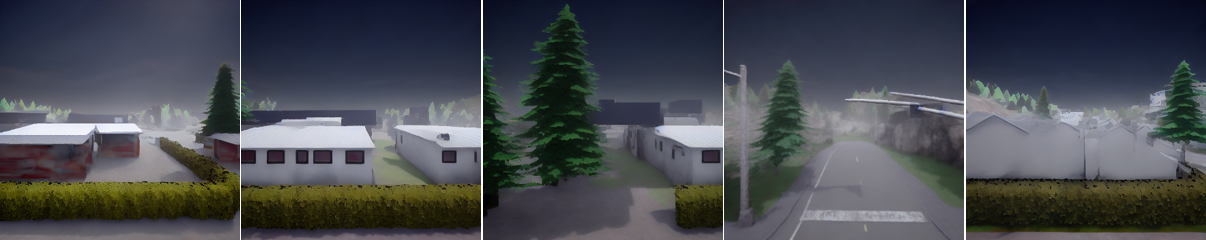}
        \end{subfigure}
    \end{minipage}

    % Fifth row
    \begin{minipage}[c]{.02\linewidth}
        \rotatebox[origin=c]{90}{Target}
    \end{minipage}%
    \hfill
    \begin{minipage}[c]{.48\linewidth}
        \begin{subfigure}[b]{\linewidth}
            \includegraphics[width=\linewidth]{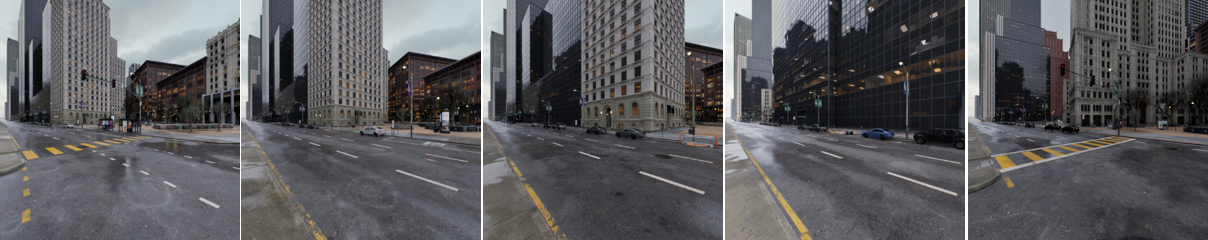}
        \end{subfigure}
    \end{minipage}%
    \hfill
    \begin{minipage}[c]{.48\linewidth}
        \begin{subfigure}[b]{\linewidth}
            \includegraphics[width=\linewidth]{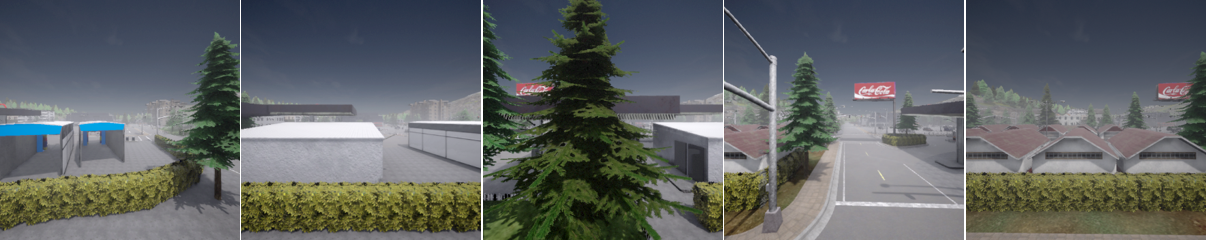}
        \end{subfigure}
    \end{minipage}
    
    \caption{\textbf{Qualitative Comparisons.} We compare Skyeyes with other SOTA methods for ground view generation. Unlike tasks that require matching ground truth, our task focuses on generating visually plausible images with continuous textures. All methods were evaluated under the same conditions, and Skyeyes consistently delivers superior visual quality.}
    \label{fig:qual_comparisons}
    \vspace{-0.5cm} % 减少caption和下方文本的间距

\end{figure*}

\paragraph{Evaluation metrics}
We use both image-based and video-wise metrics for understanding the efficacy of different techniques. For image-based evaluation, we considered metrics like PSNR~\cite{hore2010image}, SSIM~\cite{wang2004image}, LPIPS~\cite{zhang2018unreasonable} and FID~\cite{heusel2017gans}. Each of these metrics offers insights into different aspects of image quality. PSNR and SSIM are traditional measures of image quality, focusing on pixel-level accuracy and perceptual similarity, respectively. LPIPS, being a more recent metric, evaluates perceptual similarity based on learned features, providing a more nuanced understanding of visual quality. FID assesses the similarity in distribution between generated and real images, indicating the realism of the synthesized images. We also incorporated two video-wise metric, Fréchet Video Distance (FVD)~\cite{balaji2019conditional, unterthiner2019fvd} and Kernel Video Distance (KVD)~\cite{unterthiner2019fvd} to evaluate the temporal consistency and quality of video sequences.

\subsection{Results}
\begin{table}
    \caption{Quantitative comparison between Skyeyes and other state-of-the-art methods on the test set of City Sample dataset.}
    \centering
    \resizebox{\linewidth}{!}{
    \begin{tabular}{ccccccc}
        \hline
         CitySample & FID $\downarrow$ & PSNR$\uparrow$ & SSIM $\uparrow$ & LPIPS $\downarrow$ & KVD$\downarrow$ & FVD $\downarrow$ \\
         \hline
         MVS~\cite{schoenberger2016sfm} 
         & 359.15 & 27.79 & 0.30 & 0.63 & 377.20 & 2846.69 \\
         NeRF~\cite{mueller2022instant} 
         & 317.09 & 27.94 & 0.28 & 0.68 & 382.57 & 2390.31 \\
         3DGS~\cite{kerbl20233d} 
         & 245.24 & 28.13 & 0.42 & 0.62 & 340.62 & 1926.74 \\
         SuGaR~\cite{guedon2023sugar} 
         & 260.51 & 28.13 & 0.38 & 0.60 & 204.20 & 1157.64 \\
         ControlNet~\cite{zhang2023adding} & 63.47 & 28.08 & 0.25 & 0.57 & 281.89 & 1205.81 \\
         Instruct-P2P~\cite{brooks2023instructpix2pix} & 100.47 & 28.04 & 0.25 & 0.58 & 428.88 & 1742.12 \\
         GVG~\cite{xu2024geospecific} & \textbf{29.62} & 28.29 & 0.33 & \textbf{0.47} & 141.33 & 715.97\\
         Ours        
         & 54.73 & \textbf{32.22} & \textbf{0.45} & 0.48 & \textbf{117.93} & \textbf{528.65} \\
         \hline
    \end{tabular}
    }
    \label{tab:quant-region5}
    \vspace{-0.3cm}
\end{table}

\begin{table}
    \caption{Quantitative comparison between Skyeyes and other SOTA methods on the test set of CARLA dataset.}
    \centering
    \resizebox{\linewidth}{!}{
    \begin{tabular}{ccccccc}
         \hline
         CARLA & FID $\downarrow$ & PSNR$\uparrow$ & SSIM $\uparrow$ & LPIPS $\downarrow$ & KVD $\downarrow$ & FVD $\downarrow$ \\
         \hline
         MVS~\cite{schoenberger2016sfm} 
         & 388.37 & 27.82 & 0.40 & 0.53 & 562.21 & 3606.30 \\
         NeRF~\cite{mueller2022instant} 
         & 248.16 & 27.98 & 0.51 & 0.68 & 618.43 & 2571.87\\
         3DGS~\cite{kerbl20233d} 
         & 228.92 & 28.32 & 0.59 & 0.48 & 573.05 & 2404.44\\
         SuGaR~\cite{guedon2023sugar} 
         & 202.38 & 28.13 & 0.53 & 0.48 & 679.40 & 2498.16 \\
         ControlNet~\cite{zhang2023adding} & 75.26 & 27.97 & 0.58 & 0.50 & 277.89 & 1056.69 \\
         Instruct-P2P~\cite{brooks2023instructpix2pix} & 202.12 & 27.80 & 0.38 & 0.65 & 707.08 & 3327.93 \\
         GVG~\cite{xu2024geospecific} & \textbf{45.73} & 28.29 & 0.53 & 0.47 & 266.46 & 913.07 \\
         Ours        
         & 57.95 & \textbf{33.37} & \textbf{0.69} & \textbf{0.44} & \textbf{218.29} & \textbf{693.28} \\
         \hline
    \end{tabular}
    }
    \label{tab:quant-carla}
    \vspace{-0.3cm}
\end{table}

\label{sec:exp-result}
Fig.~\ref{fig:qual-fig1} visualizes the results of our proposed pipeline on two extracted datasets. Specifically, given a sequence of aerial view images, Skyeyes is able to predict and synthesize the corresponding ground view sequence.

% \paragraph{Baseline} To tackle the task of producing high-quality terrain effectively, it's crucial to investigate innovative methods. Established techniques such as Structure from Motion, NeRF and 3DGS offer promising options. Consequently, for our foundational analysis, we have chosen MVS~\cite{schoenberger2016sfm}, Instant NGP~\cite{mueller2022instant}, the original 3DGS and SuGaR~\cite{guedon2023sugar}

\paragraph{Qualitative Comparison}
We present the visual performance comparison of the baselines and our method in Figure~\ref{fig:qual_comparisons}, showcasing the effectiveness of each approach in rendering visually realistic terrain. As observed, SuGaR, while rendering geometry and color accurately, results in a somewhat blurred image, primarily due to splatting effects when viewed from this extreme perspective, especially in comparison to aerial views. ControlNet appears to produce photo-realistic images with less artifacts, their textures are significantly different from the ground truth images. Compared with GVG, our method produces less artifacts and largely maintain intra-frame consistency.

% roughly reconstructs the basic geometry but fails to capture finer details, such as building windows. NeRF exhibits anomalous coloring in the ground and lower parts of buildings, a consequence of their networks not being trained to learn features from unique viewing angles like this. Conversely, SuGaR, while rendering geometry and color accurately, results in a somewhat blurred image, primarily due to splatting effects when viewed from this extreme perspective, especially in comparison to aerial views. Our method, on the other hand, not only maintains the geometric integrity but also exhibits a degree of continuity in fine detail features.

\paragraph{Quantitative Comparison}
A significant improvement is evident in the KVD and FVD metrics, as shown in Tables~\ref{tab:quant-region5} and~\ref{tab:quant-carla}. Both KVD and FVD values are significantly lower in our method compared to others, as these metrics indicate better performance when the values are smaller. Our FVD, for instance, improves by around 25\% on average compared to the best baseline. This substantial reduction demonstrates that our method maintains superior consistency in video sequences, ensuring smooth transitions and coherence across frames. Furthermore, in terms of image-related metrics such as FID, PSNR, SSIM, and LPIPS, our method consistently ranks first or second, showcasing its competitive edge in image generation quality. In some cases, our results significantly surpass those of other methods, further emphasizing the high-quality, realistic rendering of the terrain in both image and video aspects.

% A significant improvement can be observed across all metrics in Table ~\ref{tab:quant-region5} and ~\ref{tab:quant-region5}
% %The results, as detailed in our quantitative analysis, indicate that our method outperforms the others across these metrics. 
% Notably, our method holds significantly better KVD and FVD, meaning that the generation of Skyeyes has much more similarities in the distribution of target images as well as the overall quality in longer sequence. Therefore, our method, achieving the highest PSNR and SSIM scores and the lowest LPIPS and FVD scores, demonstrates superior performance in terms of both image and video quality, reflecting enhanced realism, accuracy, and consistency in the rendered terrain.

\subsection{Ablation Study}
\label{sec:exp-ablation}
\begin{figure}[tb]
    \centering
    \begin{minipage}{\linewidth}
        \centering
        \includegraphics[width=\linewidth]{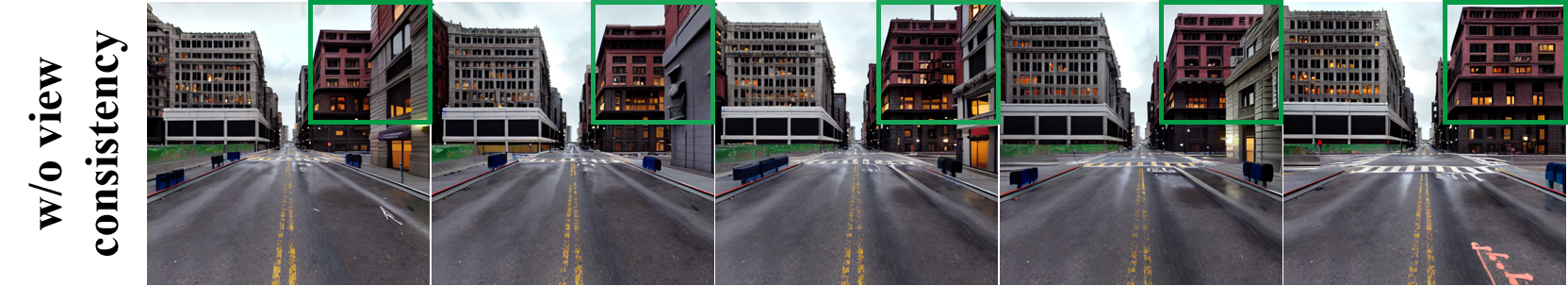}
    \end{minipage}

    \vspace{0.3cm} 

    \begin{minipage}{\linewidth}
        \centering
        \includegraphics[width=\linewidth]{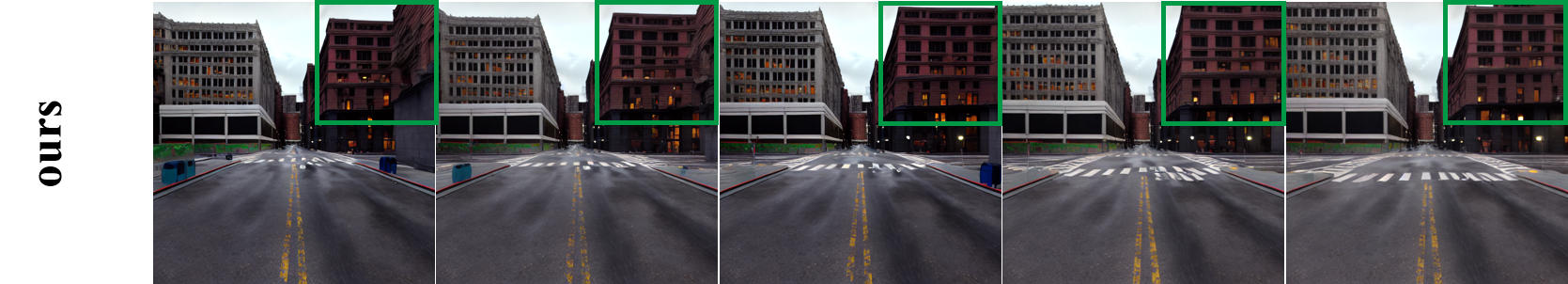}
    \end{minipage}
    
    \caption{\textbf{Ablation Study on view consistency module.} We observe apparent content inconsistency when view consistency module is dropped (top row), whereas content remain relatively consistent for our full pipeline (bottom row). The area surrounded by green square more apparently illustrates content consistency of our full pipeline.}
    \label{fig:ablation-view-consistent}
    \vspace{-10px} 

\end{figure}

\paragraph{View Consistency Module}
\label{sec:ablation-view-consistency-module}
To demonstrate the effectiveness of the View Consistency Module, we attempted to generate results by discarding the View Consistency Module while using the same prior. Top row of Fig.~\ref{fig:ablation-view-consistent} presents that if we directly use Appearance Control Module without adding the View Consistency Module, each frame produces buildings with different colors and materials. There are significant differences in the arrangement and number of windows, and even the number of tall buildings in the distance lacks continuity. With the incorporation of this module, not only is there consistency in the appearance of the buildings, but also the road markings and pedestrian crossings on the ground are somewhat consistent, greatly enhancing the continuity of the generated scene.

% \begin{figure}[tb]
%     \centering
%     \begin{minipage}{0.85\textwidth}
%         \centering
%         \includegraphics[width=\linewidth]{fig/img/experiments/ablation/seq_0040_control_full.png}
%     \end{minipage}

%     \begin{minipage}{0.85\textwidth}
%         \centering
%         \includegraphics[width=\linewidth]{fig/img/experiments/ablation/seq_0040_ours_full.png}
%     \end{minipage}
    
%     \caption{\textbf{Ablation Study on view consistency module.} We observe apparent content inconsistency when view consistency module is dropped (top row), whereas content remain relatively consistent for our full pipeline (bottom row). The area surrounded by green square more apparently illustrates content consistency of our full pipeline.}
%     \label{fig:ablation-view-consistent}
% \end{figure}

\begin{figure}[tb]
    \centering
    \begin{subfigure}{0.23\linewidth}
        \centering
        \includegraphics[width=\linewidth]{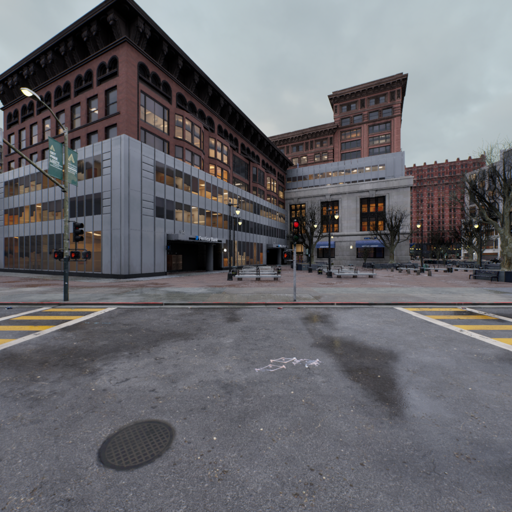}
        \caption{Ground Truth}
        % \label{fig:carla-map}
    \end{subfigure}
    \begin{subfigure}{0.23\linewidth}
        \centering
        \includegraphics[width=\linewidth]{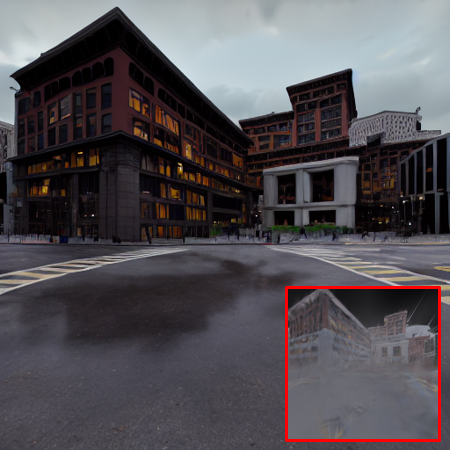}
        \caption{SuGaR~\cite{guedon2023sugar}}
        \label{fig:intro-nerf}
    \end{subfigure}
    \begin{subfigure}{0.23\linewidth}
        \centering
        \includegraphics[width=\linewidth]{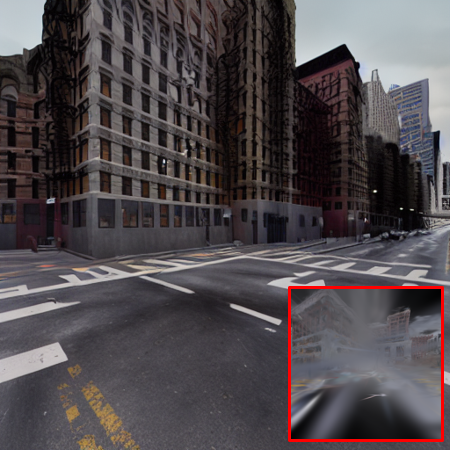}
        \caption{3DGS~\cite{kerbl20233d}}
        \label{fig:intro-3dgs}
    \end{subfigure}
    \begin{subfigure}{0.23\linewidth}
        \centering
        \includegraphics[width=\linewidth]{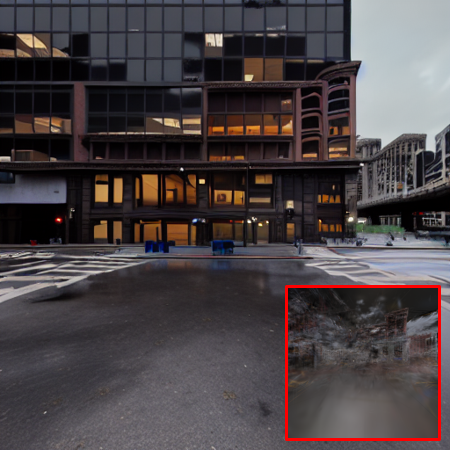}
        \caption{SC-GS~\cite{lu2024scaffold}}
        \label{fig:intro-sugar}
    \end{subfigure}

     \caption{\textbf{Ablation study on prior generation model}. We compare the generation quality based on different ground view prior. Specifically, we compare ScaffoldGS~\cite{lu2024scaffold}, 3DGS~\cite{kerbl20233d}, and SuGaR~\cite{guedon2023sugar} (ours). The red square indicates the prior generated at the same camera pose. Though experience longer training time, the prior of SuGaR presents higher generation quality.}
     \label{fig:ablation_model_choice}
     \vspace{-0.5cm}

\end{figure}

\paragraph{Different choices of ground view priors}
We study different model choices for ground view image prior generation. The ground view image prior is crucial in photo-realistic sequence generation as an ideal prior should possess more abundant content and features whereas a less ideal prior will present more blurry and noisy spaces. Therefore, we compare three different prior generation models:  3DGS~\cite{kerbl20233d}, Scaffold GS~\cite{lu2024scaffold} and SuGaR~\cite{guedon2023sugar} (our choice for the pipeline) and evaluate all of the models on CitySample dataset. The results in Fig.~\ref{fig:ablation_model_choice} indicates the generation results under the same camera pose. Although the prior of vanilla 3DGS and Scaffold GS have the potential in guiding the diffusion model to generate photo-realistic images, their appearances present significant differences compared with ground truth. In comparison, the prior of SuGaR illustrates strong ability in generation of images with controllable appearance.

% \begin{figure}[tb]
%     \centering
%     \begin{subfigure}{0.22\linewidth}
%         \centering
%         \includegraphics[width=0.98\linewidth]{fig/img/experiments/ablation/gt.png}
%         \caption{Ground Truth}
%         % \label{fig:carla-map}
%     \end{subfigure}
%     \begin{subfigure}{0.22\linewidth}
%         \centering
%         \includegraphics[width=0.98\linewidth]{fig/img/experiments/ablation/seq_0052_ours_combine.png}
%         \caption{SuGaR~\cite{guedon2023sugar}}
%         \label{fig:intro-nerf}
%     \end{subfigure}
%     \begin{subfigure}{0.22\linewidth}
%         \centering
%         \includegraphics[width=0.98\linewidth]{fig/img/experiments/ablation/seq_0052_skyeyes_combine.png}
%         \caption{3DGS~\cite{kerbl20233d}}
%         \label{fig:intro-3dgs}
%     \end{subfigure}
%     \begin{subfigure}{0.22\linewidth}
%         \centering
%         \includegraphics[width=0.98\linewidth]{fig/img/experiments/ablation/seq_0052_nerf_combine.png}
%         \caption{NeRF~\cite{mueller2022instant}}
%         \label{fig:intro-sugar}
%     \end{subfigure}
    
%      \caption{\textbf{Ablation study on prior generation model}. We compare the generation quality based on different ground view prior. Specifically, we compare NeRF~\cite{mueller2022instant}, 3DGS~\cite{kerbl20233d} and SuGaR~\cite{guedon2023sugar} (ours). The red square indicates the prior generated at the same camera pose. Though experience longer training time, prior of SuGaR presents higher generation quality.}
%      \label{fig:ablation_model_choice}
% \end{figure}

\section{Limitations and Future Work}
\label{sec:limitation}

% While SkyEyes marks a significant advancement in aerial-to-ground cross-view synthesis, there are certain limitations that need to be addressed in future work:

% \begin{itemize}
%     \item \textbf{Generalization to Real-World Data:} 
    One primary limitations of Skyeyes is its current performance in generalizing to real-world data. The framework, as it stands, is largely trained on synthetic datasets extracted from simulators like City Sample. While these datasets offer a controlled environment for training, they may not fully capture the complexity and variability found in real-world scenarios. The textures, lighting conditions, and architectural elements in synthetic environments can differ significantly from those in real-world settings. This discrepancy can lead to challenges in achieving the same level of detail and realism when the model is applied to actual aerial and ground-level imagery. Addressing this limitation will involve refining the training datasets to include more diverse and realistic scenarios.
%     \item \textbf{Lack of an End-to-End Training Process:} Another limitation of our current methodology is the absence of an end-to-end training process. The framework operates in multiple stages, each requiring distinct processing steps. The initial phase involves training the 3D Gaussian Splatting model, followed by the application of appearance and view consistency modules for enhancing photo-realism and maintaining temporal coherence. While this modular approach allows for specific focus on different aspects of the terrain generation process, it also introduces complexity and potential inefficiencies. An end-to-end training process could streamline these steps, leading to a more cohesive and efficient generation pipeline. 
% \end{itemize}
\section{Conclusion}

Our research introduces Skyeyes, a groundbreaking framework designed for aerial-to-ground cross-view synthesis, adeptly transforming aerial imagery into detailed and realistic 3D terrain models. This innovative approach, a first in large-scale outdoor scene generation, skillfully integrates 3DGS with controllable diffusion models. This integration not only identifies and fills data gaps but also provides robust prior feature to the controllable generation of diffusion model. The diffusion model further enhances this framework by ensuring noise control and maintaining spatiotemporal consistency, thereby producing superior quality results compared to traditional video-to-video synthesis methods.
Our experimental results demonstrate Skyeyes' effectiveness in creating high-quality, realistic terrain models. This success is evident in the framework's ability to surpass existing methods in terms of visual accuracy and consistency. Skyeyes stands out as a significant advancement in terrain generation.

\section{Acknowledgement}
Supported by the Intelligence Advanced Research Projects Activity (IARPA) via Department of Interior/ Interior Business Center (DOI/IBC) contract number 140D0423C0075. The U.S. Government is authorized to reproduce and distribute reprints for Governmental purposes notwithstanding any copyright annotation thereon. Disclaimer: The views and conclusions contained herein are those of the authors and should not be interpreted as necessarily representing the official policies or endorsements, either expressed or implied, of IARPA, DOI/IBC, or the U.S. Government.

{\small
\bibliographystyle{ieee_fullname}
\bibliography{PaperForReview}
}

\end{document}


\title{Supplementary Materials - Skyeyes: Ground Roaming using Aerial View Images}

\author{
Zhiyuan Gao\textsuperscript{1,2,\thanks{Equal Contribution}}\quad Wenbin Teng\textsuperscript{1,2,\footnotemark[1]}\quad Gonglin Chen\textsuperscript{1,2}\quad Jinsen Wu\textsuperscript{1,2}\\
Ningli Xu\textsuperscript{3} \quad Rongjun Qin\textsuperscript{3} \quad Andrew Feng\textsuperscript{2} \quad Yajie Zhao\textsuperscript{1,2,\thanks{Corresponding Author}}\\\\
\textsuperscript{1}University of Southern California\quad \textsuperscript{2}Institute for Creative Technologies\quad \textsuperscript{3}The Ohio State University\\
\ttfamily\small \{gaozhiyu, wenbinte, gonglinc, jinsenwu\}@usc.edu\\
\ttfamily\small \{xu.3961\}@buckeyemail.osu.edu \quad \{Qin.324\}@osu.edu \quad \{feng, zhao\}@ict.usc.edu
}

\maketitle

% Task:
% 1. Done - 写一下baseline为什么选择这几个 - scott
% 2. Done - 图5 保留两个 - scott
% 3. Done - 图6 说明目的不是重建不是生成 - scott
% 4. 图6 增加Instruct pix2pix/controlnet+每行5张图 - wenbin
% 5. 表1 表2把数据补全（增加三行）- wenbin
% 6. Done - 缩减到8页 - scott
% 7. 图8，增加scaffold和2dgs - wenbin
% 8. Done - Baseline描述符合表的情况 - scott
% 9. Done- supplementary整理 - Scott

% 1. 数据集收集的详细信息
% 2. 更多的visual results from skyeyes
% 3. Qualitative Comparisons from other methods
% 4. videos

In this supplementary material, we provide more details of our dataset collection in Section~\ref{sec:supp_dataset}. After that, we provide additional qualitative result in Section~\ref{sec:supp_additional} and additional ablation studies in Section~\ref{sec:supp_additional_ablation}. In addition, based on the limitation of this work introduced in the main paper, we discuss our potential future work in Section~\ref{sec:supp_future_work}.

\section{Dataset Collection}
\label{sec:supp_dataset}
\subsection{CARLA Simulator}
\begin{figure*}
    \centering
    \begin{subfigure}{0.6\textwidth}
        \centering
        \includegraphics[width=\linewidth]{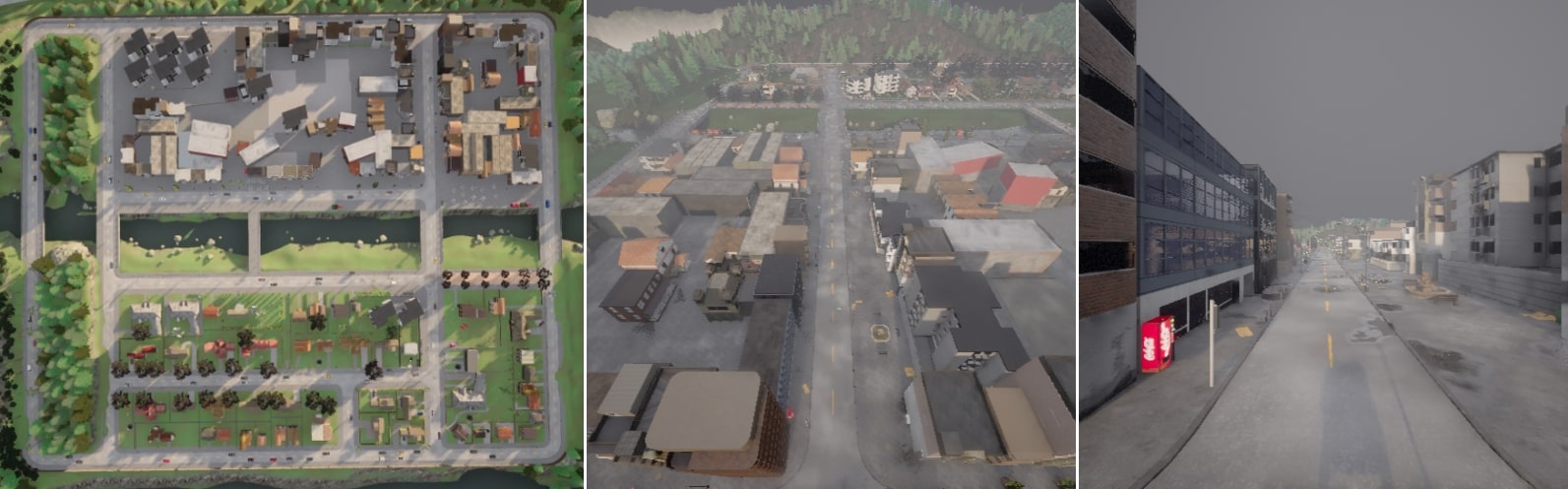}
    \end{subfigure}
    \begin{subfigure}{0.6\textwidth}
        \centering
        \includegraphics[width=\linewidth]{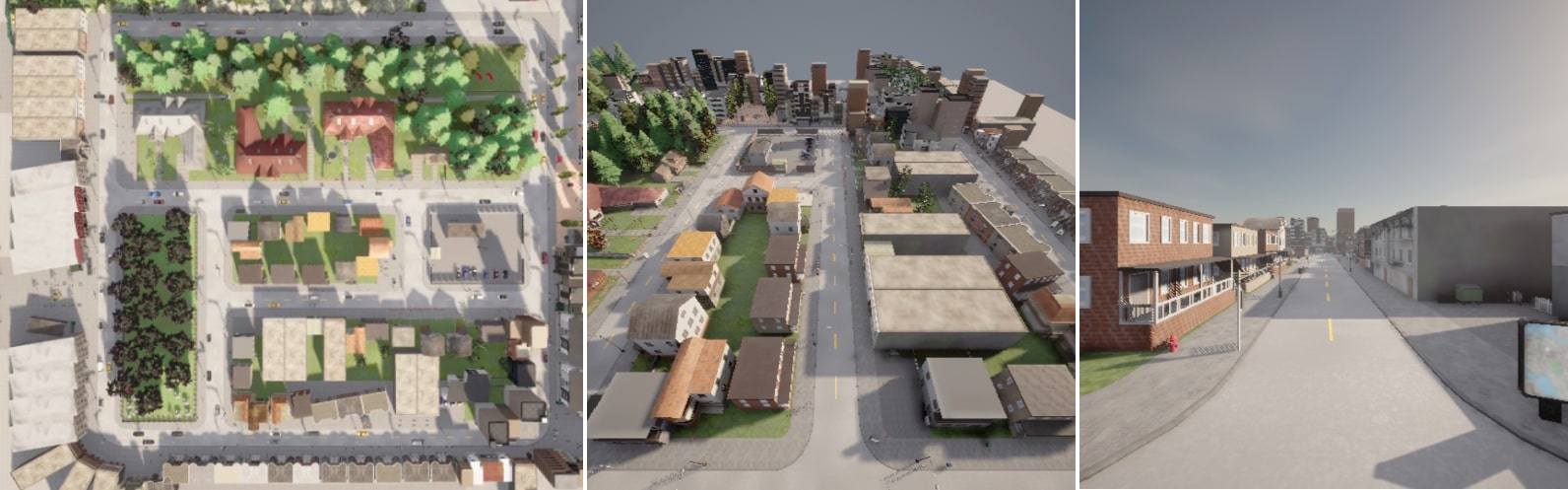}
    \end{subfigure}
    \begin{subfigure}{0.6\textwidth}
        \centering
        \includegraphics[width=\linewidth]{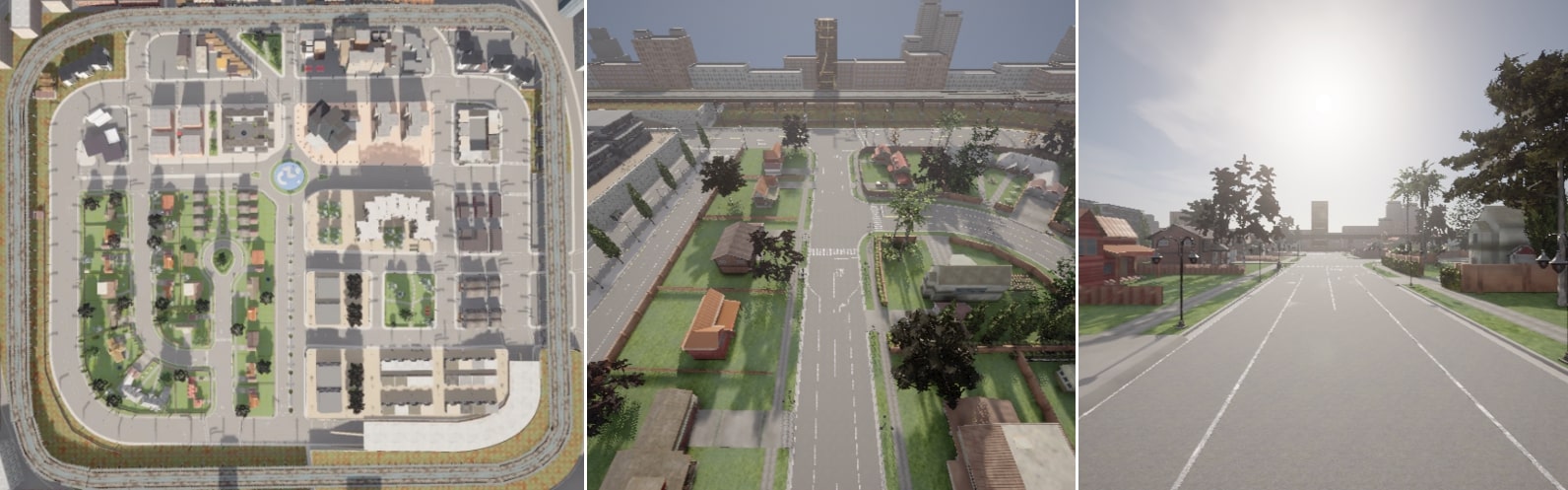}
    \end{subfigure}
    \begin{subfigure}{0.6\textwidth}
        \centering
        \includegraphics[width=\linewidth]{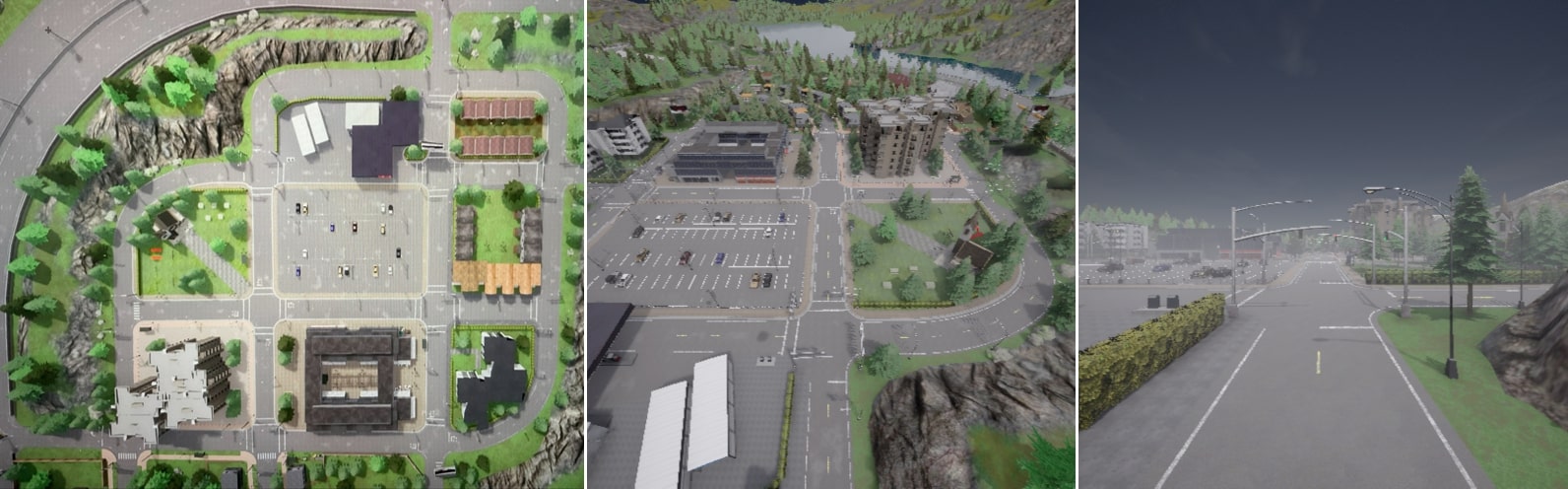}
    \end{subfigure}
    \begin{subfigure}{0.6\textwidth}
        \centering
        \includegraphics[width=\linewidth]{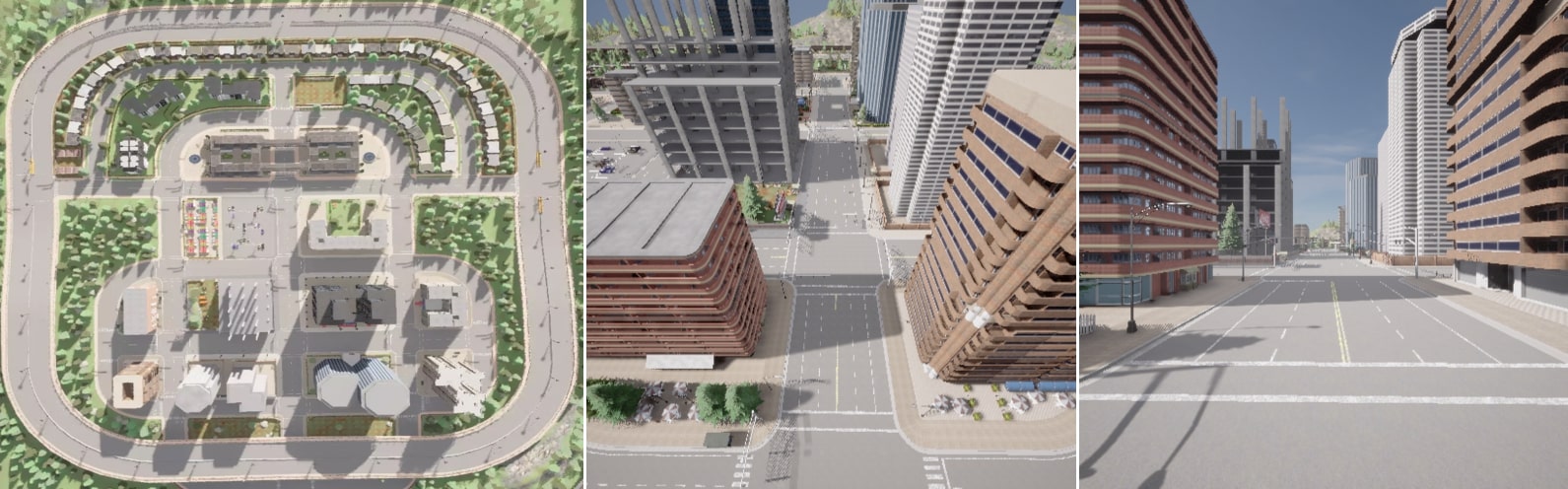}
    \end{subfigure}
    \begin{subfigure}{0.6\textwidth}
        \centering
        \includegraphics[width=\linewidth]{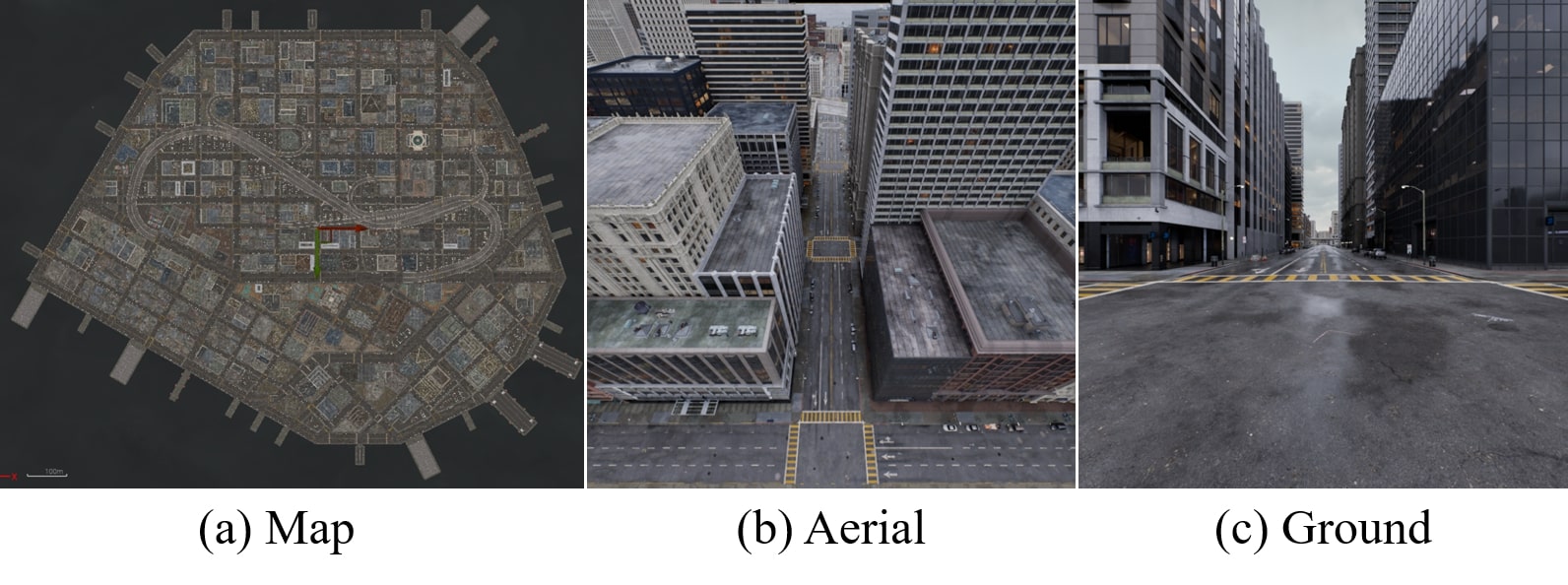}
    \end{subfigure}
    
    \caption{\textbf{Dataset Visualization.} The first four rows are Town01, Town02, Town03, Town04 and Town05 of CARLA Simulator~\cite{Dosovitskiy17}, respectively. The last row is a visualization of CitySample dataset.}
    \label{fig:supp_dataset}
\end{figure*}

The CARLA Simulator~\cite{Dosovitskiy17} provides comprehensive Python API to facilitate interactions between users and environment. We leverage the Python API to build connections with the CarlaUE4 server, load the target map, add ego vehicle and multiple sensor cameras. We design customized trajectories for the vehicle and render the whole scene with sensor cameras. The first 4 rows of Figure~\ref{fig:supp_dataset} illustrates the top down view of each town that we extract data from together with an example of aerial/ground pairs. For more examples, please see our supplementary video. We separate each scene with multiple lanes. Within each lane, we spawn cameras to capture a color image for every 2 meters. Camera positioning is automatically optimized by CARLA to adhere to constraints like maintaining a safe distance from buildings, and the camera orientation is adjusted to face the direction of travel. For each point sampled on the lane, we set the yaw value of camera rotation to vary within $k\pi/4$, where $k=0...7$. The altitude of aerial sequence is set to be 52 meters while the altitude of ground sequence is 2 meters. Pitch value of camera rotation is set to be -45 degrees for aerial views whereas 0 for ground views. For training-evaluation purposes, we set all the extracted data from Town04 for evaluation and all the rest for training. 

\subsection{CitySample}
As discussed in the main paper, we follow the same data extraction pipeline as MatrixCity~\cite{li2023matrixcity}. We only manipulate the rotation and position of camera trajectories to extract our customized data. Similarly, please refer to the last row of Figure~\ref{fig:supp_dataset} and our supplementary video for examples of CitySample dataset. The data extraction protocol mirrors the data extraction strategy employed in the CARLA Simulator, where the starting and ending points of each lane were manually determined. The configuration of camera poses in this environment closely aligns with those in CARLA, with the notable distinction that aerial sequences are captured at an altitude of 100 meters. We choose region 5 as the test set, and region 1, 2, 3, 4 and 6 as the training set  (See Figure 3 in our main paper)

\section{Additional Visualization}
\label{sec:supp_additional}
\subsection{Additional Results}
\begin{figure*}
    \centering
    \begin{subfigure}{0.85\textwidth}
        \centering
        \includegraphics[width=\linewidth]{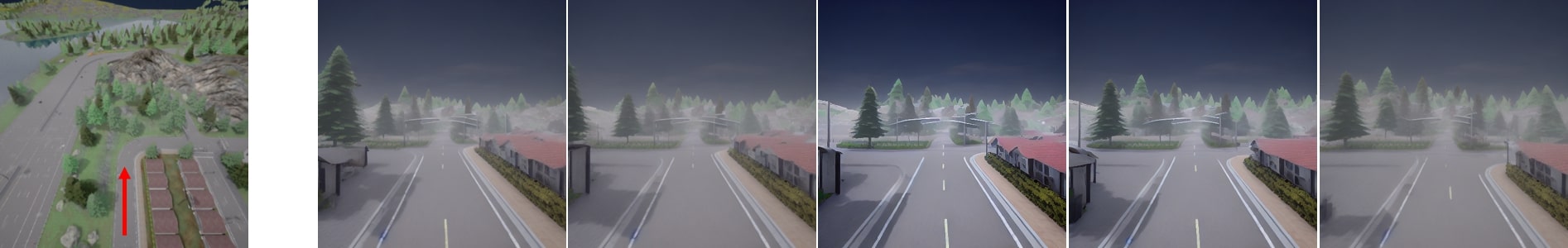}
    \end{subfigure}
    \begin{subfigure}{0.85\textwidth}
        \centering
        \includegraphics[width=\linewidth]{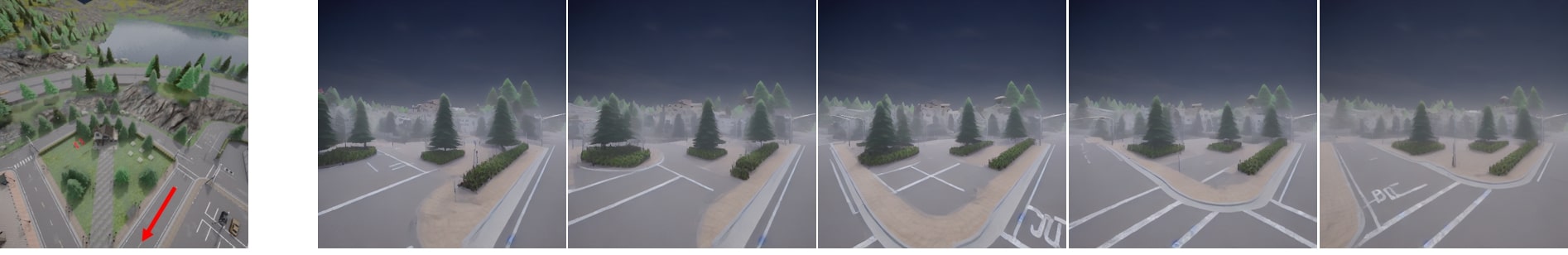}
    \end{subfigure}
    \begin{subfigure}{0.85\textwidth}
        \centering
        \includegraphics[width=\linewidth]{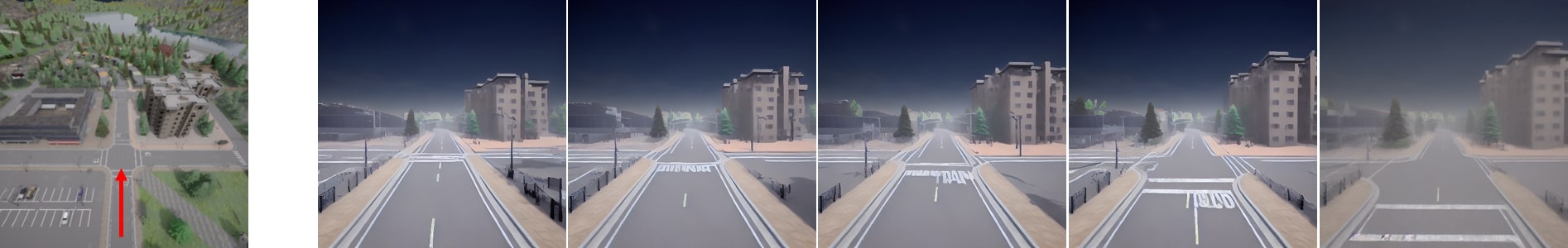}
    \end{subfigure}
    \begin{subfigure}{0.85\textwidth}
        \centering
        \includegraphics[width=\linewidth]{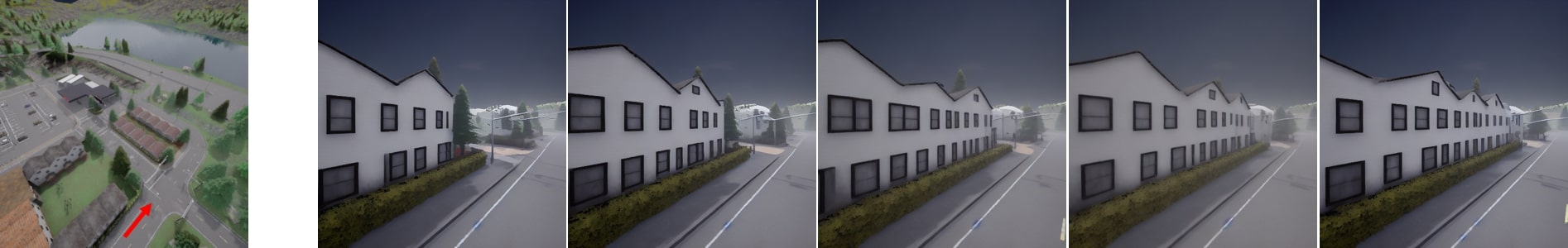}
    \end{subfigure}
    \begin{subfigure}{0.85\textwidth}
        \centering
        \includegraphics[width=\linewidth]{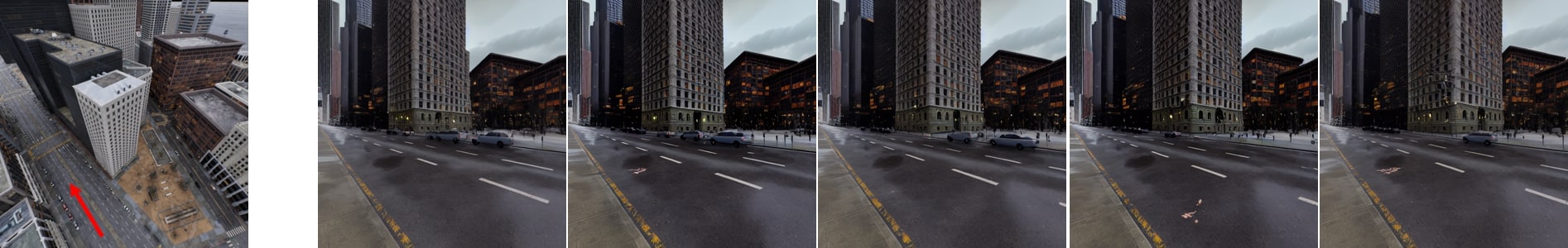}
    \end{subfigure}
    \begin{subfigure}{0.85\textwidth}
        \centering
        \includegraphics[width=\linewidth]{fig/img/supplement/add_result/citysample_1.jpg}
    \end{subfigure}
    \begin{subfigure}{0.85\textwidth}
        \centering
        \includegraphics[width=\linewidth]{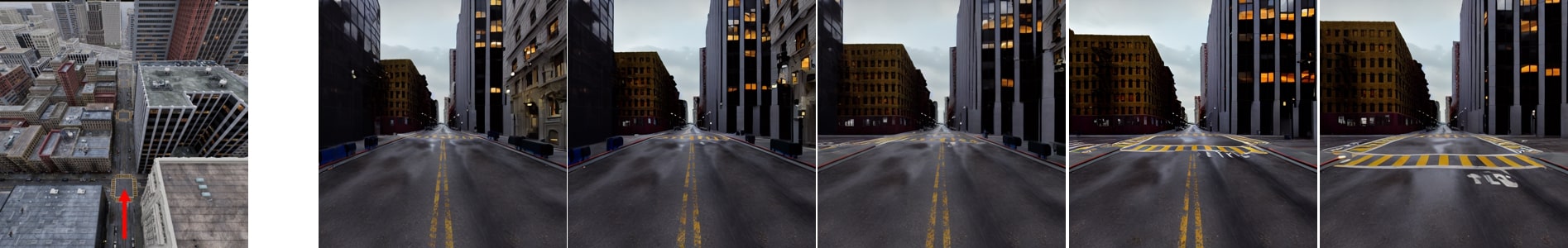}
    \end{subfigure}
    \begin{subfigure}{0.85\textwidth}
        \centering
        \includegraphics[width=\linewidth]{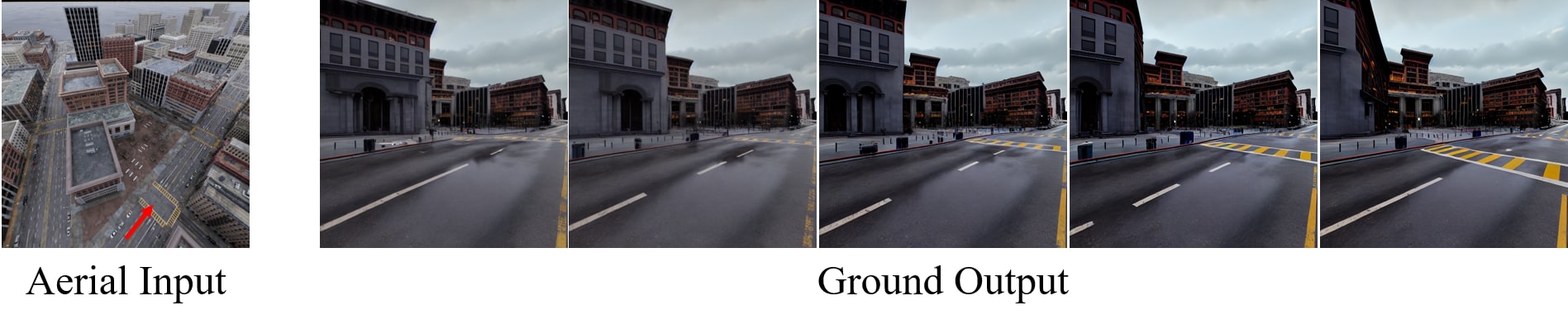}
    \end{subfigure}
    
    \caption{Additional visualizations of aerial view to ground view synthesis on CARLA and CitySample datasets}
    \label{fig:supp_add_results}
\end{figure*}

Figure~\ref{fig:supp_add_results} provides visualization of addition evaluation of our method on the test set of CARLA and CitySample dataset

\subsection{Long Video Generation}
As discussed in the main paper, we first use appearance control module to generate the first image, which is further applied as a condition to generate the following frames. For longer video generation, we simply use the last frame generated from current sequence as the new condition to generate next sequence. Please refer to our supplementary video for long video generation results.

\section{Additional Ablation}
\label{sec:supp_additional_ablation}
\begin{figure*}
    \centering
    \begin{subfigure}{0.85\textwidth}
        \centering
        \includegraphics[width=\linewidth]{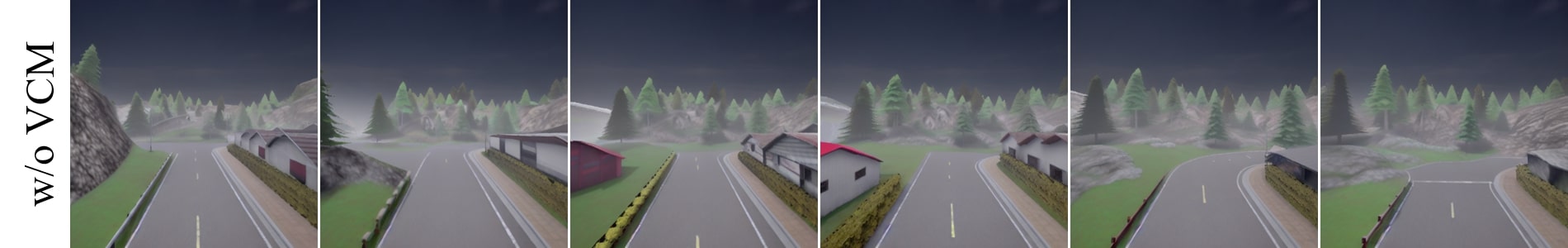}
    \end{subfigure}
    \begin{subfigure}{0.85\textwidth}
        \centering
        \includegraphics[width=\linewidth]{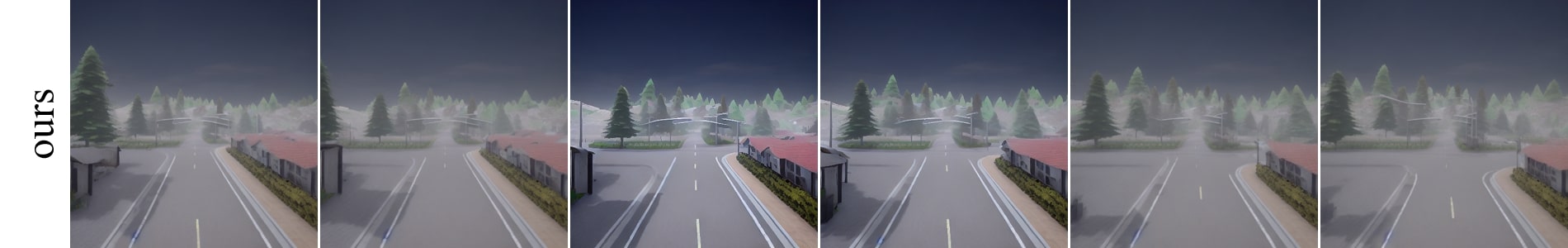}
    \end{subfigure}
    \begin{subfigure}{0.85\textwidth}
        \centering
        \includegraphics[width=\linewidth]{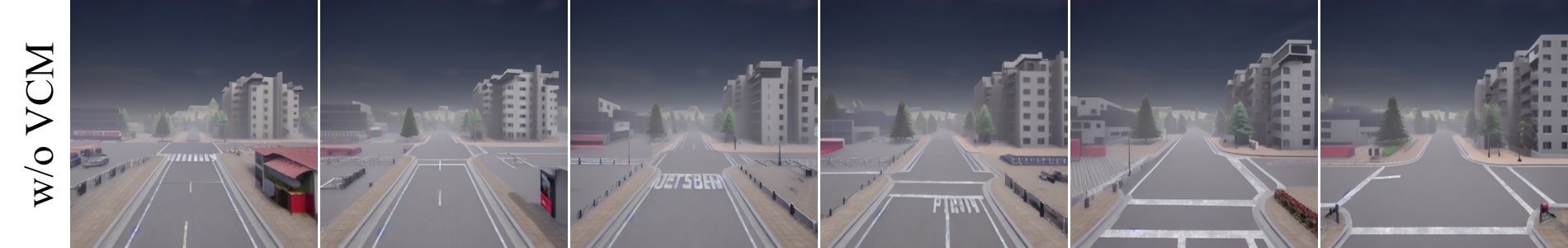}
    \end{subfigure}
    \begin{subfigure}{0.85\textwidth}
        \centering
        \includegraphics[width=\linewidth]{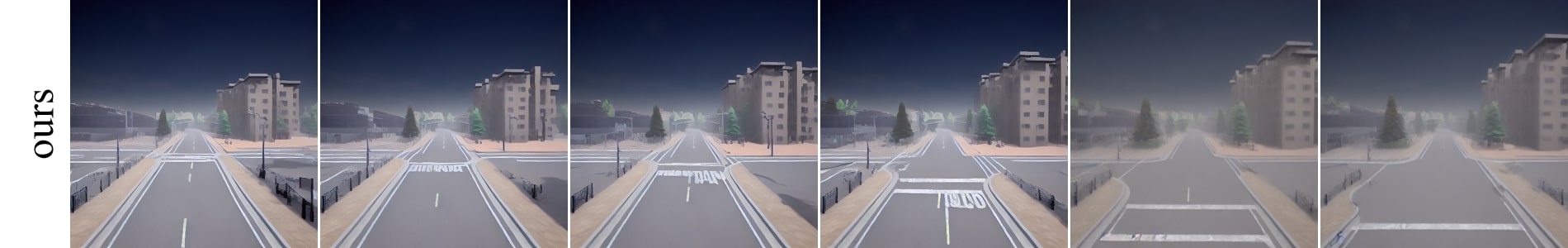}
    \end{subfigure}
    \begin{subfigure}{0.85\textwidth}
        \centering
        \includegraphics[width=\linewidth]{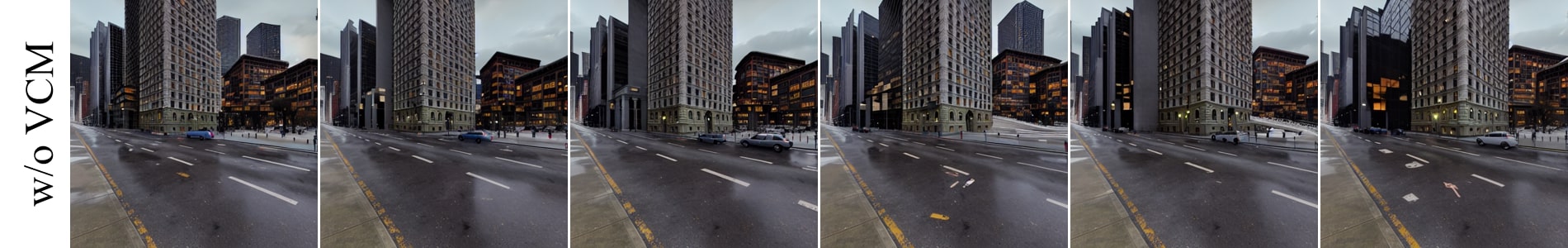}
    \end{subfigure}
    \begin{subfigure}{0.85\textwidth}
        \centering
        \includegraphics[width=\linewidth]{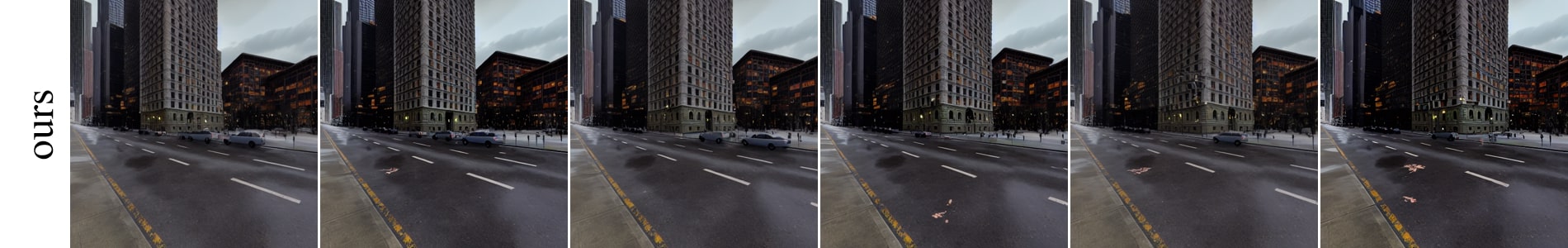}
    \end{subfigure}
    \begin{subfigure}{0.85\textwidth}
        \centering
        \includegraphics[width=\linewidth]{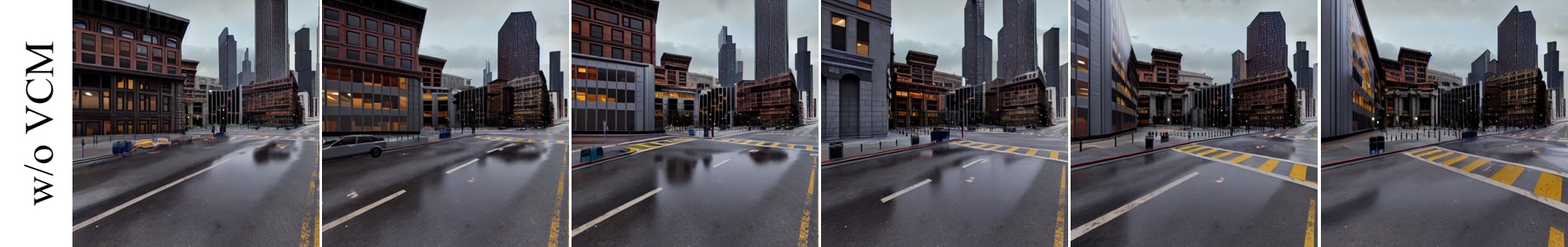}
    \end{subfigure}
    \begin{subfigure}{0.85\textwidth}
        \centering
        \includegraphics[width=\linewidth]{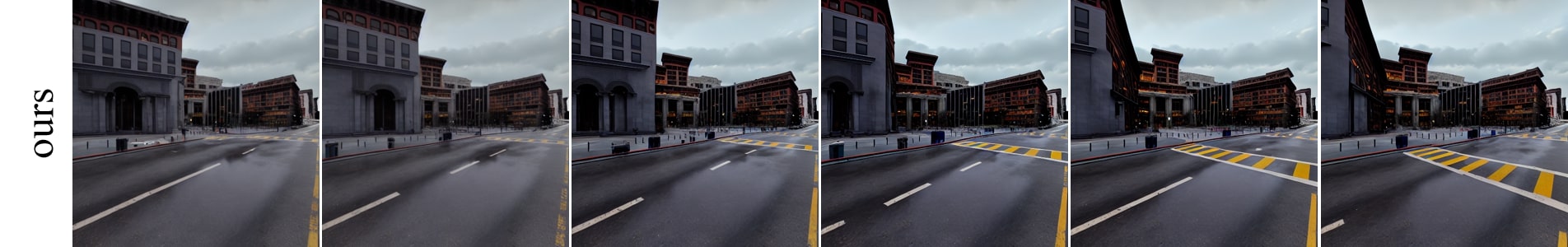}
    \end{subfigure}
    
    \caption{Additional ablation studies on view consistency module (VCM). For better visualizations, we pick 6 nearby positions along 4 different ground view sequences and display generation results for without and with VCM in every other rows.}
    \label{fig:supp_add_ablation}
\end{figure*}

Figure~\ref{fig:supp_add_ablation} provides visualization of ablation experiments of view consistency module. Also, please refer to our supplementary video for more detailed visualizations.

\section{Discussion and Future Work}
\label{sec:supp_future_work}
As discussed in the main paper, our proposed method does not generalize well to the realistic data. This is mainly due to the lack of scale and variety of the training data, which is currently limited to synthetic data with two open source platforms. However, the extraction of large amount of geo-aligned aerial-to-ground pairwise data is very costly and the acquisition should abide by the local rules and policies. Therefore, our next step is to perform unsupervised domain adaptation in diffusion model to bridge the gap between synthetic and realistic dataset.


{\small
\bibliographystyle{ieee_fullname}
\bibliography{PaperForReview}
}